\documentclass[11pt]{article}

\usepackage[final]{acl}

\usepackage{times}
\usepackage{latexsym}

\usepackage[T1]{fontenc}

\usepackage[utf8]{inputenc}

\usepackage{microtype}

\usepackage{inconsolata}

\usepackage{graphicx}

\usepackage{algorithm}
\usepackage{algorithmic}

\usepackage{booktabs}
\usepackage{amssymb}
\usepackage{amsmath}

\usepackage{amsfonts}
\usepackage{textcomp}
\usepackage{xcolor}
\usepackage{colortbl}

\definecolor{Gray}{gray}{0.9}

\usepackage{multirow}
\usepackage{subcaption}
\usepackage{arydshln}

\usepackage{newfloat}
\usepackage{listings}

\newcommand{\ourmethodfirst}{ImCoref}
\newcommand{\ourmethod}{ImCoref-CeS}
\definecolor{Gray}{gray}{0.9}

\title{ImCoref-CeS: An \underline{Im}proved Lightweight Pipeline for \underline{Coref}erence Resolution with LLM-based \underline{C}heck\underline{e}r-\underline{S}plitter Refinement}

\author{\textbf{Kangyang Luo$^{\spadesuit}$, Yuzhuo Bai$^{\spadesuit}$, Shuzheng Si$^{\spadesuit}$, Cheng Gao$^\spadesuit$, Zhitong Wang$^{\spadesuit}$}\\
\textbf{Yingli Shen$^\spadesuit$, Wenhao Li$^\spadesuit$, Zhu Liu$^{\spadesuit}$, Yufeng Han$^{\spadesuit}$, Jiayi Wu$^\diamondsuit$, Cunliang Kong$^\spadesuit$} \\ 
\textbf{Maosong Sun\thanks{Corresponding author}$^{\spadesuit}$$^\clubsuit$$^\bigstar$ }\\
$^{\spadesuit}$Department of Computer Science and Technology, Tsinghua University 
\\ $^\clubsuit$Institute for AI, Tsinghua University \quad $^\diamondsuit$East China Normal University \\ $^\bigstar$Jiangsu Collaborative Innovation Center for Language Ability}

\begin{document}
\maketitle
\begin{abstract}
Coreference Resolution (CR) is a critical task in Natural Language Processing (NLP). Current research faces a key dilemma: whether to further explore the potential of supervised neural methods based on small language models, whose detect-then-cluster pipeline still delivers top performance, or embrace the powerful capabilities of Large Language Models (LLMs). However, effectively combining their strengths remains underexplored. To this end, we propose \textbf{ImCoref-CeS}, a novel framework that integrates an enhanced supervised model with LLM-based reasoning. First, we present an improved CR method (\textbf{ImCoref}) to push the performance boundaries of the supervised neural method by introducing a lightweight bridging module to enhance long-text encoding capability, devising a biaffine scorer to comprehensively capture positional information, and invoking a hybrid mention regularization to improve training efficiency. Importantly, we employ an LLM acting as a multi-role Checker-Splitter agent to validate candidate mentions (filtering out invalid ones) and coreference results (splitting  erroneous clusters) predicted by ImCoref. Extensive experiments demonstrate the effectiveness of ImCoref-CeS, which achieves superior performance compared to existing state-of-the-art (SOTA) methods.

\end{abstract}

\section{Introduction}
\begin{figure}[t]
  \centering
  \includegraphics[width=0.8\linewidth]{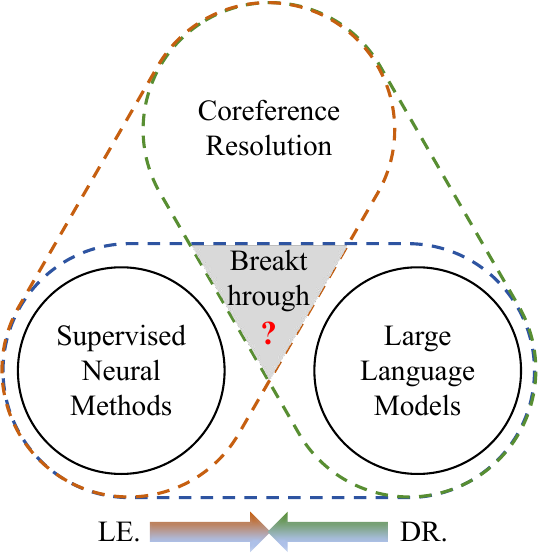}
  \caption{Combining the strengths of supervised neural methods and LLMs for CR, where  LE. and DR. are \textit{low-resource efficiency} and \textit{deep reasoning}, respectively.} 
  \label{pic_intro:}
\end{figure}
Coreference resolution (CR) aims to detect and cluster distinct text spans (referred to as mentions) within a document that refer to the same entity~\cite{karttunen1976discourse, lee2017end, lee2018higher}. As a fundamental component, CR plays a critical role in various downstream applications, including text summarization~\cite{liu2024bridging}, knowledge graph construction~\cite{pan2024unifying}, question answering~\cite{pan2024unifying, jang-etal-2025-ambiguity}, and named entity recognition~\cite{shang2025local}. Current mainstream CR methods can broadly fall into three types: \textbf{supervised neural methods}~\cite{xu-choi-2020-revealing, wu-etal-2020-corefqa, kirstain2021coreference, lai2022end, otmazgin2022f, otmazgin2022lingmess, martinelli-etal-2024-maverick} built upon small pre-trained language models (e.g., SpanBERT~\cite{joshi2020spanbert}, DeBERTa~\cite{he2021deberta}), \textbf{generative methods}~\cite{liu-etal-2022-autoregressive, bohnet-etal-2023-coreference, zhang-etal-2023-seq2seq} based on sequence-to-sequence architectures (e.g., mT5xxl~\cite{xue-etal-2021-mt5}, T0-11B~\cite{ivison2022hint}), and \textbf{LLMs}~\cite{le2023large, gan-etal-2024-assessing, zhu-etal-2025-llmlink} leveraging zero-shot learning (e.g., GPT-4~\cite{achiam2023gpt}). Given the constraints of academic computational resources, generative methods are often not the preferred choice due to their high training costs and inference latency. Notably, recent supervised neural model Maverick$_{\text{mes}}$~\cite{martinelli-etal-2024-maverick} trained with DeBERTa has achieved SOTA performance. In contrast, while LLMs struggle with the mention detection, preventing their coreference performance from surpassing supervised neural methods, they show a significant strength: given gold mentions, LLMs with powerful reasoning capabilities can achieve competitive coreference results~\cite{le2023large}. However, this prerequisite is rarely met in practical scenarios.

Therefore, existing research indicates that in CR, the SOTA supervised neural methods demonstrate notable advantages in terms of training cost, inference efficiency, and coreference performance, with their potential still not yet fully tapped. However, owing to limited model scale and task-specific nature, they tend to generate excessive invalid mentions on out-of-domain data~\cite{toshniwal2021generalization, xia-van-durme-2021-moving}. Furthermore, even on in-domain data, the presence of invalid mentions and coreference errors impedes further performance gains~\cite{martinelli-etal-2024-maverick}. Consequently, a pivotal question arises (see Fig.~\ref{pic_intro:}): \textit{\textbf{Can the reasoning capabilities of LLMs be effectively employed to address these inherent limitations of supervised neural methods?}}

This paper presents an affirmative solution by proposing a novel CR framework, coined \textbf{ImCoref-CeS}, which further explores the performance potential of the supervised neural method under constrained computational resources, and synergistically integrates the strengths of LLMs (powerful reasoning capabilities) and supervised neural methods (low-resource efficiency) during inference. Specifically, we first formulate an improved CR method (called \textbf{ImCoref}) by extending the existing leading detect-then-cluster pipeline architecture (i.e., Maverick) with three key refinements: a lightweight bridging module (LBM) is introduced to enhance long-text encoding capability; a biaffine scorer is designed  to better capture position information of end tokens during training; a hybrid mention regularization strategy is proposed to improve training efficiency.

Building upon this, we then dynamically incorporate the LLM as a multi-role Checker-Splitter agent into the inference pipeline via meticulously engineered prompts. During mention detection, ImCoref~generates candidate mentions, which are then one-by-one validated by the LLM checker based on the local context, with invalid mentions being removed. During mention clustering, ImCoref~clusters the validated mentions, thereby yielding initial coreference results. The LLM first checks coreference clusters with at least two mentions. For clusters deemed incorrect (i.e., not all mentions refer to the same entity), the LLM splitter regroups the mentions to ensure intra-cluster coreference consistency and inter-cluster exclusivity. Notably, to balance performance and resource consumption, mention and coreference cluster filtering mechanisms are introduced via direct matching and probability ranking.

We conducted extensive experiments on multiple CR benchmarks. The results demonstrate that ImCoref~achieves consistent improvements over current SOTA supervised neural methods w.r.t. coreference performance. Importantly, the ImCoref-CeS~framework advances beyond these results, enabling higher levels of performance.

\begin{figure*}[t]
  \centering
  \includegraphics[width=1.0\linewidth]{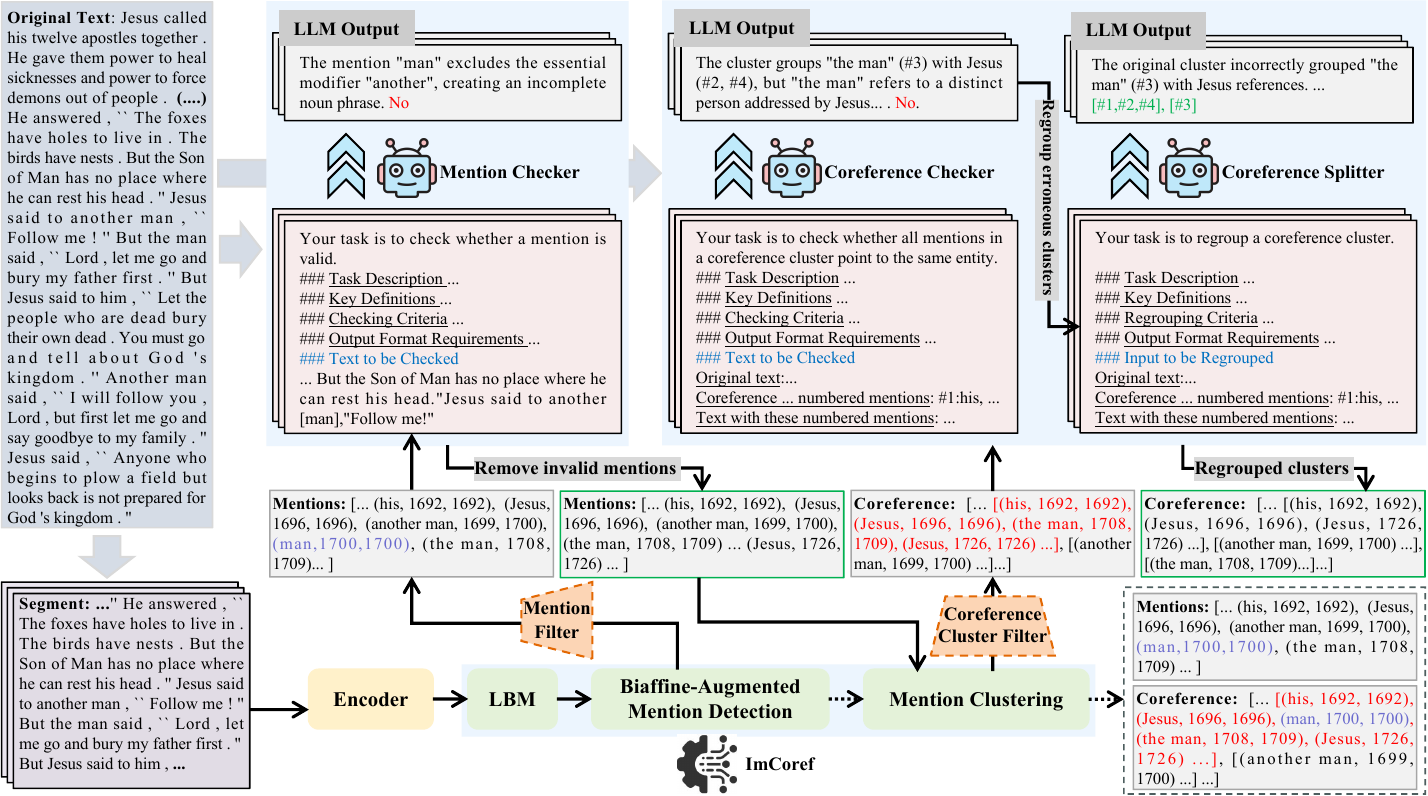}
  \caption{The overall pipeline of \ourmethod. The dashed arrow indicates the process path where only ImCoref~is executed; its generated mentions and coreference results are enclosed within the dashed box. It can be observed that relying solely on \ourmethodfirst~during inference may produce invalid mentions (e.g., {\ttfamily\small (man,1700,1700)}). Furthermore, \ourmethodfirst~inherits these invalid mentions, propagating them into erroneous coreference results. To mitigate these issues, we introduce a LLM as a multi-role Checker-Splitter agent, dynamically integrating it with \ourmethodfirst.}
  \label{Framework_fig:}
\end{figure*}

\section{Preliminaries}
\textbf{Notations.} 
CR aims to identify different mentions that point to the same entity and cluster them into coreferential chains within a given long text $D=[S_1, S_2, \cdots, S_n]$, which consists of $n$ segments; each segment $S_i=[t_{i1}, t_{i2},\cdots,t_{im_{i}}]$ contains $m_{i}$ tokens.  Also, $\mathbf{H}_i=(\mathbf{h}_{i1}, \cdots, \mathbf{h}_{im_i}) \in \mathbb{R}^{m_i\times d_h}$ is the hidden representation of $S_i$ from the text encoder, where $d_h$ is the hidden dimension size of the text encoder. And let $M$ be the number of tokens contained in $D$. Then, the hidden representation of $D=[t_1,\cdots,t_M]$ is $\mathbf{H}=(\mathbf{h}_1, \cdots, \mathbf{h}_M) \in \mathbb{R}^{M\times d_h}$. 

\textbf{Maverick.} To the best of our knowledge, Maverick~\cite{martinelli-etal-2024-maverick} is the SOTA neural CR method under resource-constrained setting. 
It employs a two-step scheme: \textit{\textbf{predict start positions of mentions first, then conditionally predict their end positions}.}
A mention start probability $p_{start}(\mathbf{H})=\sigma( \mathrm{MLP}(\mathbf{H}))$ is computed for each token, with $\sigma$ denoting the sigmoid function and $\mathrm{MLP}$ representing a multilayer perceptron classifier.
For tokens $t_s$ with $p_{start}(\mathbf{H})\succeq 0.5$ (i.e., with representation $\mathbf{H}_s$), the method conditionally predicts the probability of a subsequent token $t_e\geq t_s$ being the corresponding end position as:
\begin{align}
    p_{end}(\mathbf{H}_e|\mathbf{H}_s)= \sigma(\mathrm{MLP}(\tilde{\mathbf{H}}_{s} \oplus \mathbf{H}_{e})), \label{end_prob_mlp_eq:}
\end{align}
where $\tilde{\mathbf{H}}_{s}$ is the duplicated start token representation aligned with all possible end representations $\mathbf{H}_{e}$, and $\oplus$ denotes concatenation operator. Candidate mentions with $p_{end}(\mathbf{H}_e|\mathbf{H}_s)\succeq 0.5$ are passed to the mention clustering stage.



\section{Methodology}
In this section, we detail our proposed framework \ourmethod, including \textit{long-text encoding with lightweight
bridging module (LBM)}, \textit{biaffine-augmented mention detection}, \textit{hybrid mention regularization} and \textit{LLM Checker-Splitter}. 

\subsection{Long-Text Encoding with LBM}
Within CR, long texts are common. Existing methods~\cite{joshi2019bert, toshniwal-etal-2020-learning, guo-etal-2023-dual} typically segment these texts utilize strategies like \textit{independent} or \textit{overlapping} (see Fig.~\ref{pic2_lbm:app} in Appendix~\ref{app:full_pronouns_list}), and then sequentially concatenate the hidden representations. 
However, they isolate the semantics of different segments during both training and inference, hindering the model's ability to capture long-distance coreference relations\footnote{\textit{overlapping}  mitigates semantic isolation to some extent, but it remains incomplete.}~\cite{martinelli-etal-2024-maverick}. 
To tackle this shortcoming, we introduce a lightweight Bridging Module (\textbf{LBM}) during the sequential concatenation of hidden representations, which is designed to propagate semantic information from preceding segments to subsequent ones. 

Specifically, 
when encoding each segment $S_i$, we prepend a \texttt{[CLS]} token and append a \texttt{[SEP]} token, forming $\mathbf{H}_i=(\mathbf{h}_{i[cls]}, \mathbf{h}_{i1}, \cdots, \mathbf{h}_{im_i}, \mathbf{h}_{i[sep]})$ $\in \mathbb{R}^{(m_i+2)\times d_h}$.
To establish semantic links between adjacent segments $S_i$ and $S_{i+1}$, LBM processes the hidden representation of \texttt{[SEP]} token from $S_i$ (i.e., $\mathbf{h}_{i[sep]}$), along with that of segment $S_{i+1}$ (i.e., $\mathbf{H}_{i+1}$), as follows:
\begin{align}
    \mathbf{H}_{i+1} \leftarrow \text{LBM}\left( \mathbf{h}_{i[sep]}, \mathbf{H}_{i+1} \right).
\end{align}
Here, $\mathbf{h}_{i[sep]}$ serves as the holistic semantic representation of segment $S_i$. LBM uses this representation to modulate $\mathbf{H}_{i+1}$, enabling contextual continuity across segments through left-to-right propagation.
In our experiments, we implement two variants of LBM: 
\begin{align} 
    \hat{\mathbf{H}}_{i+1}&=\left\{\begin{array}{l}
            \text{FC}( \text{Ex}(\mathbf{h}_{i[sep]}) \oplus \mathbf{H}_{i+1} ) \\
            \text{MHA}( \mathbf{h}_{i[sep]}, \mathbf{H}_{i+1} )
            \end{array}  \right. ,\\
    \mathbf{H}_{i+1} &\leftarrow \text{Norm}( \hat{\mathbf{H}}_{i+1} + \mathbf{H}_{i+1} ), 
\end{align}
where FC and MHA are fully connected and multi-head attention layers, respectively. Ex(·) operator broadcasts $\mathbf{h}_{i[sep]}$ to match the dimensions of $\mathbf{H}_{i+1}$,  and Norm(·) denotes standard layer normalization. 
The specific choice of LBM implementation is flexible. Also, LBM is employed exclusively on the \textit{independent} strategy in our paper (see Fig.~\ref{pic2_lbm:app}).

\subsection{Biaffine-Augmented Mention Detection}
We revisit and observe that Maverick underutilizes positional information during calculating the conditional probabilities of end positions, i.e., Eq.(\ref{end_prob_mlp_eq:}). To address this, we propose a biaffine scorer inspired by~\citet{dozat2016deep, xu2022multi, mao2024span} to enhance positional modeling. 
Concretely, it integrates the start hidden representations $\tilde{\mathbf{H}}_s$ with the end that $\mathbf{H}_e$ by the following form:
\begin{align}
    & \mathbf{X}_s = \mathrm{FC}_{1}(\tilde{\mathbf{H}}_s), \mathbf{X}_e = \mathrm{FC}_{2}(\mathbf{H}_e), \\ 
    & \mathbf{S} = \mathbf{X}_{s}^T \mathbf{U} \mathbf{X}_{e} + \mathbf{W}(\mathbf{X}_{s} \oplus \mathbf{X}_{e}) + \mathbf{b}, 
\end{align}
where $\mathbf{U}\in \mathbb{R}^{d_h\times d_r \times d_h}$, $\mathbf{W}\in \mathbb{R}^{d_r \times 2d_h}$ and $b\in \mathbb{R}^{d_r}$ are learnable parameters ($d_r$ denotes the output dimension). Also, $\mathrm{FC}_{k}(\cdot):\mathbb{R}^{d_h}\rightarrow \mathbb{R}^{d_h}(k\in [2])$. 
Then, the $p_{end}(\mathbf{H}_e|\mathbf{H}_s)$ is computed as:
\begin{align}
    p_{end}(\mathbf{H}_e|\mathbf{H}_s)= \sigma(\mathrm{MLP}(\mathbf{S})).\label{biaffline_eq11:}
\end{align}




\subsection{Hybrid Mention Regularization}
Moreover, during mention detection, the EOS mention regularization strategy in the Maverick method aims to reduce computational cost, but it may induce the model to predict overly long invalid mentions.
To mitigate this issue, we propose a novel hybrid mention regularization strategy (abbreviated as HyMR).
To be specific, for the start position $t_s$ of a mention, the mention's possible end positions $t_e$ ($t_e \geq t_s$) must satisfy $t_e - t_s \leq L_s = \min \{L_{\text{max}}, L_{\text{EOS}_s}\}$, where $L_{\text{max}}$ is the predefined maximum span length, and $L_{\text{EOS}_s}$ is the distance from $t_s$ to the nearest EOS token. 
Notably, $L_{\text{max}}$ explicitly restricts excessively long spans in lengthy sentences, while $L_{\text{EOS}_s}$ ensures that mentions do not cross sentence boundaries, thus mitigating biases that may arise from relying on a single constraint, such as generating overly long invalid mentions or disrupting sentence structures.
In contrast, $L_s = L_{\text{EOS}_s}$ in the Maverick method. 
Therefore,  HyMR intuitively further lowers the training cost.

So far, \textit{\textbf{we have described our improvements to current SOTA supervised neural method}}, with the method termed \textbf{\ourmethodfirst}. In our experiments, the mention clustering strategy of Maverick$_\text{mes}$ is adopted by \ourmethodfirst.


\subsection{LLM Checker-Splitter}
Our method, \ourmethod, augments \ourmethodfirst~with a dynamic LLM-based Checker-Splitter agent. This is achieved through two key designs: 1) integration of the LLM via carefully engineered prompts, and 2) the introduction of mention and coreference cluster filters based on direct matching and probability ranking. 
The complete pipeline is depicted in Fig.~\ref{Framework_fig:}.

\textbf{Checker-Splitter}. 
The Checker-Splitter agent leverages the LLM to refine the outputs of \ourmethodfirst~through two sequential steps: mention validation and cluster verification/splitting.
During \textit{\textbf{mention detection}}, the module validates candidate mentions produced by \ourmethodfirst. 
The validity of each mention is judged by the LLM (acting as a \textit{mention checker}) after it is annotated within its local context---typically its host sentence and a limited preceding context.
During \textit{\textbf{mention clustering}}, the initial coreference clusters from \ourmethodfirst~are verified by the LLM acting as a \textit{coreference checker}. 
If a cluster is identified as erroneous (i.e., its mentions do not corefer), the LLM acts as a \textit{coreference splitter} to partition the mentions into mutually exclusive, coreferentially consistent groups. The context provided for these operations is strategically limited to sentences relevant to the cluster.

To guide the LLM in performing the above checking and splitting functions, we design structured prompt templates (see Fig.~\ref{Framework_fig:}), each comprising \textit{system instructions}, \textit{task description}, \textit{key definitions}, \textit{checking/regrouping criteria}, and \textit{output format requirements}. 
Due to space limitations, the detailed annotation guidelines for mentions and coreference clusters, along with the complete templates and examples, are provided in Appendices~\ref{annotation_guid:app} and~\ref{app:prompt_temp}, respectively.
\textit{\textbf{Notably, while a full-text clustering fusion strategy based on the splitter's outputs could potentially enhance performance, we note that its implementation presents non-trivial challenges and that our current improvements already achieve substantial gains.}} We therefore leave this exploration for future work.


\textbf{Mention and Coreference Cluster Filters.} 
To balance performance and resource cost, we introduce mention and coreference cluster filters, restricting LLM calls to only those mentions or clusters within long texts that require checking or splitting.
The specific mechanisms are as follows:

The \textit{\textbf{mention filter}} first employs direct matching to bypass pronoun checks (e.g., {\ttfamily\small (his,1692,1692)}; see the full pronoun list in Appendix~\ref{app:full_pronouns_list}). This strategy is based on the linguistic prior~\cite{otmazgin2022lingmess, martinelli-etal-2024-maverick} that pronouns (with few exceptions such as ``it'') generally refer to entities, and are limited in number and form, thus allowing efficient filtering. Subsequently, for non-pronoun mentions, we select those with probability $p_{end}(\mathbf{H}_e|\mathbf{H}_s)\succeq 0.5$, rank them in descending order of probability, and use the LLM to validate only the bottom $\eta_1$ of them.
The \textit{\textbf{coreference cluster filter}} applies two screening conditions: (1) only initial clusters containing more than one mention are considered; and (2) for each multi-mention cluster $c$, a ranking metric  $p^f_c$ is constructed based on the pairwise clustering probabilities from \ourmethodfirst, defined as:
\begin{align}
    p^f_c = \overline{p}_c - \rho \sum_{j=1}^{|c|} (\overline{p}_c - p_{c,j})^2,
\end{align}
where $p_{c,j}$ denotes the $j$-th pairwise clustering probability in cluster $c$, $|c|$ is the total number of such probability pairs, $\overline{p}_c = \frac{1}{|c|} \sum_{j=1}^{|c|} p_{c,j}$ is the average probability, and $\rho$ is a sufficiently small positive real number. 
The metric $ p^f_c$, dominated by $\overline{p}_c$ due to small $\rho$, serves as a confidence score for coreference. We therefore rank clusters by descending $p^f_c$ and submit only the bottom $\eta_2$ of clusters (i.e., the least confident ones) to the LLM for verification and splitting.

The above probability ranking mechanism is inspired by~\cite{chen-etal-2024-double}: when \ourmethodfirst~assigns a lower predictive probability to a mention or coreference cluster, it indicates higher uncertainty about its correctness, thus warranting greater need for LLM intervention.

\section{Experiments}

\subsection{Experimental Settings}

\textbf{Datasets.} We train and evaluate the performance of our proposed methods on two widely used CR datasets: OntoNotes~\cite{pradhan-etal-2012-conll} and LitBank~\cite{bamman-etal-2020-annotated}. 
To assess the generalization capability of methods, models trained on OntoNotes are also tested on the out-of-domain dataset WikiCoref~\cite{ghaddar-langlais-2016-wikicoref}. 
We detail these datasets in Table~\ref{tab:Dataset_statistics}.

\begin{table}[htbp]
  \centering
  \resizebox{1.0\columnwidth}{!}{
    \begin{tabular}{l|ccccc}
    \hline
    Datasets & \#Train & \#Val & \#Test & Avg. W & Avg. M  \\
    \hline
    OntoNotes & 2802  & 343   & 348   & 467   & 56      \\
    LitBank & 80    & 10    & 10    & 2105  & 291     \\
    WikiCoref & -     & -     & 30    & 1996  & 230    \\
    \hline
    \end{tabular}%
    }
    \caption{Dataset statistics: number of documents in each dataset split (\#Train/Val/Test), average number of words (i.e., Avg. W) and mentions (i.e., Avg. M) per document.}
  \label{tab:Dataset_statistics}%
  \vspace*{-2ex}
\end{table}%

\textbf{Baselines.} To gauge the effectiveness of our methods, we follow~\citet{martinelli-etal-2024-maverick} by adopting DeBERTa$_{\text{large}}$~\cite{he2021deberta} as the base model for \ourmethodfirst. 
Meanwhile, we choose three latest and most robust LLMs for \ourmethod: Qwen2.5-72B-Instruct~\cite{qwen2.5}, DeepSeek-R1-250528~\cite{deepseekai2025deepseekr1incentivizingreasoningcapability} and GPT-4~\cite{le2023large}. For clarity, we denote \ourmethod~with different LLMs as \textbf{\ourmethod$_{\text{qwen/dpsk/gpt4}}$}. 
Also, we compare \ourmethod~and \ourmethodfirst~against numerous baselines: 1) \textit{\textbf{Supervised neural methods}} including c2f-coref~\cite{joshi2020spanbert}, ICoref~\cite{xia-etal-2020-incremental}, CorefQA~\cite{wu-etal-2020-corefqa}, s2e-coref~\cite{kirstain2021coreference}, longdoc~\cite{toshniwal2021generalization}, wl-coref~\cite{dobrovolskii-2021-word}, f-coref~\cite{otmazgin2022f}, LingMess~\cite{otmazgin2022lingmess}, Dual-cache~\cite{guo-etal-2023-dual}, and Maverick$_{\text{mes}}$~\cite{martinelli-etal-2024-maverick}; 2) \textit{\textbf{Generative methods}} containing ASP~\cite{liu-etal-2022-autoregressive}, Link-Append~\cite{bohnet-etal-2023-coreference}, and seq2seq~\cite{zhang-etal-2023-seq2seq}; 3) \textit{\textbf{Large Language Models}} comprising InstructGPT~\cite{le2023large}, GPT-3.5-turbo~\cite{le2023large, zhu-etal-2025-llmlink}, GPT-4~\cite{le2023large}, and LLMLink~\cite{zhu-etal-2025-llmlink}.  
Notably, for fairness, our methods align with the text encoding strategy of Maverick$_{\text{mes}}$ in comparative experiments, without considering the LBM module. The utility of LBM will be thoroughly investigated in the ablation study.

\begin{table*}[htbp]
  \centering
  \resizebox{1.95\columnwidth}{!}{
    \begin{tabular}{llccccccl|cc}
    \hline
    \multicolumn{1}{c}{\multirow{2}[4]{*}{Methods}} & \multicolumn{1}{c}{\multirow{2}[4]{*}{Base Encoders}} & \multirow{2}[4]{*}{MUC  } & \multirow{2}[4]{*}{B$^3$} & \multirow{2}[4]{*}{ CEAF$_{\phi_4}$} & \multirow{2}[4]{*}{Avg. F1} & \multirow{2}[4]{*}{Params} & \multicolumn{2}{c|}{Training} & \multicolumn{2}{c}{Inference} \\
\cline{8-11}          &       &       &       &       &       &       & Time  & \multicolumn{1}{c|}{Hardware} & Time  & Mem. \\
    \hline
    \rowcolor{Gray}\multicolumn{11}{l}{\textbf{Supervised neural methods}} \\
    c2f-coref & SpanBERT$_\text{large}$ & 85.3  & 78.1  & 75.3  & 79.6  & 370M  & -     & \multicolumn{1}{r|}{1$\times$ 32G} & 50s   & 11.90 \\
    Icoref & SpanBERT$_\text{large}$ & 85.3  & 77.8  & 75.2  & 79.4  & 377M  & 40h   & \multicolumn{1}{r|}{1$\times$1080TI-12G} & 38s   & 2.90 \\
    CorefQA & SpanBERT$_\text{large}$ & 88.0  & 82.2  & 79.1  & 83.1$^\diamond$  & 740M  & -     & \multicolumn{1}{r|}{1$\times$TPUv3-128G} & -     & - \\
    s2e-coref & LongFormer$_\text{large}$ & 85.8  & 79.1  & 76.1  & 80.3  & 494M  & -     & \multicolumn{1}{r|}{1$\times$ 32G} & 17s   & 3.90 \\
    longdoc & LongFormer$_\text{large}$ & 85.3  & 78.0  & 75.3  & 79.6  & 471M  & 16h   & \multicolumn{1}{r|}{1$\times$ A6000-48G} & 25s   & 2.10 \\
    wl-coref & RoBERTa$_\text{large}$ & 86.3  & 79.9  & 76.6  & 81.0  & 360M  & 5h    & \multicolumn{1}{r|}{1$\times$RTX8000-48G} & 11s   & 2.30 \\
    f-coref & DistilRoBERTa & 84.4  & 76.6  & 74.5  & 78.5$^\diamond$  & 91M   & -     & \multicolumn{1}{r|}{1$\times$V100-32G} & 3s    & 1.00 \\
    LingMess & LongFormer$_\text{large}$ & 86.6  & 80.5  & 77.3  & 81.4  & 590M  & 23h   & \multicolumn{1}{r|}{1$\times$V100-32G} & 20s   & 4.80 \\
    Dual cache & LongFormer$_\text{large}$ & 86.3  & 80.3  & 76.8  & 81.1  & 471M  & -     & \multicolumn{1}{r|}{1$\times$T4-16G} & -     & - \\
    Maverick$_\text{mes}$ & DeBERTa$_\text{large}$ & 88.0  & 82.8  & 79.9  & 83.6  & 504M  & 14h   & \multicolumn{1}{r|}{1$\times$RTX4090-24G} & 14s   & 4.00 \\
    ImCoref(\textbf{ours}) & DeBERTa$_\text{large}$ & \textbf{88.2}  & \textbf{83.6}  & \textbf{81.0}  & \textbf{84.3}  & 531M  & 13h   & \multicolumn{1}{r|}{1$\times$RTX4090-24G} & 15s   & 4.05 \\
    \hline
    \rowcolor{Gray}\multicolumn{11}{l}{\textbf{Generative methods}} \\
    ASP   & FLAN-T5$_\text{XXL}$ & 87.2  & 81.7  & 78.6  & 82.5  & 11B   & 45h   & \multicolumn{1}{r|}{6$\times$A100-80G} & 20m   & - \\
    Link-Append & mT5$_\text{XXL}$ & \textbf{87.8}  & \textbf{82.6}  & \textbf{79.5}  & \textbf{83.3}  & 13B   & 48h   & \multicolumn{1}{r|}{128$\times$TPUv4-32G} & 30m   & - \\
    seq2seq & T0$_\text{XXL}$ & 87.6  & 82.4  & 79.5  & 83.2  & 11B   & -     & \multicolumn{1}{r|}{8$\times$A100-80G} & 40m   & - \\
    \hline
    \rowcolor{Gray}\multicolumn{11}{l}{\textbf{Large Language Models}} \\
    \multirow{2}[0]{*}{InstructGPT} & \multirow{2}[0]{*}{API} & 70.4  & 58.4  & 51.7  & 60.1  & -     & -     & \multicolumn{1}{c|}{-} & -     & - \\
          &       & 89.2  & 79.4  & 73.7  & 80.8$^\star$  & -     & -     & \multicolumn{1}{c|}{-} & -     & - \\
    \hdashline
    \multirow{2}[0]{*}{GPT-3.5-turbo} & \multirow{2}[0]{*}{API} & 66.9  & 55.5  & 46.5  & 56.3  & -     & -     & \multicolumn{1}{c|}{-} & -     & - \\
          &       & 86.2  & 79.3  & 68.3  & 77.9$^\star$  & -     & -     & \multicolumn{1}{c|}{-} & -     & - \\
    \hdashline
    \multirow{2}[1]{*}{GPT-4} & \multirow{2}[1]{*}{API} & 73.7  & 62.7  & 52.3  & 62.9  & -     & -     & \multicolumn{1}{c|}{-} & -     & - \\
          &       & \textbf{93.7}  & \textbf{88.8}  & \textbf{82.8}  & \textbf{88.4}$^\star$  & -     & -     & \multicolumn{1}{c|}{-} & -     & - \\
    \hline
    \rowcolor{Gray}\multicolumn{11}{l}{\textbf{Supervised neural methods+Large Language Models (ours)}} \\
    ImCoref-CeS$_\text{qwen}$ & DeBERTa$_\text{large}$+API & 89.6  & 84.1  & 81.0  & 84.9  & 531M  & 13h   & 1 $\times$ RTX4090-24G & 3m    & 4.05 \\
    ImCoref-CeS$_\text{dpsk}$ & DeBERTa$_\text{large}$+API & 90.8  & 84.6  & 81.6  & 85.7  & 531M  & 13h   & 1 $\times$ RTX4090-24G & 5m    & 4.05 \\
    ImCoref-CeS$_\text{gpt4}$ & DeBERTa$_\text{large}$+API & \textbf{91.2}  & \textbf{84.9}  & \textbf{81.8 } & \textbf{86.0}  & 531M  & 13h   & 1 $\times$ RTX4090-24G & 2m    & 4.05 \\
    \hline
    \end{tabular}%
    }
  \caption{
   Results of different methods on OntoNotes. We report the MUC, B$^3$, and CEAF$_{\phi_4}$ F1 scores (\%), their average (Avg. F1), base encoders, model parameters (Params), training time/hardware, and inference time/memory (Mem., GiB). ($\diamond$) indicates models trained with additional resources; ($\star$) signifies clustering was performed using gold mentions. Full performance comparisons are provided Table~\ref{tab:Per_com_ontonotes_app} in Appendix~\ref{app:add_exp_res}.
  }
  \label{tab:Per_com_ontonotes}%
  \vspace*{-2ex}
\end{table*}

\textbf{Metrics.}
In our experiments, coreference performance across different methods is evaluated using the MUC~\cite{vilain1995model}, B$^3$~\cite{bagga1998algorithms}, and CEAF$_{\phi_4}$~\cite{luo2005coreference} metrics, with Precision (P), Recall (R), and F1 scores reported for each. The overall performance of a method is measured by the CoNLL-F1 score, computed as the average of the F1 scores from these three metrics (abbreviated as Avg.F1). To further analyze model capabilities, certain ablation studies also report Precision (P), Recall (R), and F1 scores for Mention Detection. 

\textbf{Configurations.}
Unless otherwise specified, we default to setting $L_{\text{max}}=30$ for HyMR, $\eta = \eta_1=\eta_2=0.6$, and $\rho=1e-3$. Meanwhile, we employ the Adafactor optimizer~\cite{shazeer2018adafactor} for model training, setting the learning rate for the DeBERTa$_{\text{large}}$ to 2e-5 and for the remaining model layers to 3e-4. 
All experiments are implemented using the PyTorch-Lightning framework. Each run is executed on a single RTX 4090 GPU with 24GB of VRAM.  
For the LLMs utilized, we access them remotely via their API interfaces, uniformly setting the temperature parameter to $0$.
Due to space limitations, the complete experimental settings are provided in Appendix~\ref{app:com_exp_set}.

\subsection{Results Comparison}

\textbf{OntoNotes.} We compare the proposed methods against the current mainstream CR baselines on the OntoNotes dataset, with the results presented in Table~\ref{tab:Per_com_ontonotes}. 
The results show that: 
\textbf{1)} \ourmethod~outperforms all existing mainstream methods in terms of Avg.F1, with \ourmethod$_{\text{gpt4}}$ achieving the best coreference performance of 86.0\%. This confirms that integrating LLMs as multi-role Checker-Splitter agents during inference effectively enhances the coreference capability of \ourmethodfirst. Of note, while GPT-4 reaches an Avg. F1 of 88.4\% under the gold mention setting, its performance severely deteriorates without this precondition, lagging notably behind all existing supervised neural methods. This result corroborates our previous discussion on the limitations of LLMs in mention detection. 
\textbf{2)} \ourmethodfirst~surpasses all other supervised neural methods w.r.t. Avg.F1, notably surpassing Maverick$_{\text{mes}}$ on most metrics. From Table~\ref{tab:Per_com_ontonotes}, although \ourmethodfirst~has slightly more parameters (531M vs. 504M), it requires less training time (13h vs. 14h) under the same hardware setup, owing to the efficiency of the HyMR strategy. Moreover, both models exhibit comparable inference speed and memory usage. These results indicate that the proposed biaffine scorer and the HyMR effectively augment mention detection, leading to improved coreference resolution without significantly increasing computational costs.
\textbf{3)} Comparing \ourmethodfirst~and \ourmethod, we find that while the latter improves coreference performance, it introduces additional inference latency. Nevertheless, compared to generative approaches such as ASP and Link-Append, which require dozens of minutes per document, \ourmethod~maintains a practical inference time (2--5 minutes). Therefore, we recommend using \ourmethod~in scenarios demanding high accuracy but not requiring real-time response, and adopting \ourmethodfirst~when faster inference is critical. It should be noted that API-based LLM access makes inference time network-dependent.
Importantly, we present a qualitative error analysis in Appendix~\ref{app:error_ana}.

\begin{table}[htbp]
  \centering
  \resizebox{1.0\columnwidth}{!}{
    \begin{tabular}{lccc|c}
    \hline
    \multicolumn{1}{c}{Methods} & MUC   & B$^3$ & CEAF$_{\phi_4}$ & Avg.F1 \\
    \hline
    longdoc & \textbf{88.2}  & 75.9  & 65.5  & 76.5  \\
    Dual-cache & \textbf{88.2}  & 79.2  & 71.0  & 79.5  \\
    Maverick$_{\text{mes}}$ & 86.5  & 78.8  & 69.8  & 78.4  \\
    ImCoref(\textbf{ours}) & 87.9  & \textbf{79.5}  & \textbf{71.6}  & \textbf{79.7}  \\
    \hline
    seq2seq & -     & -     & -     & 77.3 \\
    \hline
    GPT-3.5-turbo & -     & -     & -     & 75.3  \\
    LLMLink & -     & -     & -     & 81.5$^\star$  \\
    \hline
    ImCoref-CeS$_{\text{qwen}}$ & 88.7  & 80.2  & 72.3  & 80.4 \\
    ImCoref-CeS$_{\text{dpsk}}$ & \textbf{89.1}  & \textbf{81.9}  & \textbf{74.5}  & \textbf{81.8}  \\
    ImCoref-CeS$_{\text{gpt4}}$ & 88.8  & 81.5  & 72.9  & 81.1  \\
    \hline
    \end{tabular}%
    }
    \caption{Performance comparison (\%) of different methods on LitBank. Note that ($\star$)  indicates that it is required to train multiple LLMs to perform coreference task.}
  \label{tab:Per_com_litbank}%
  \vspace*{-2ex}
\end{table}%

\textbf{LitBank.}
We compare the proposed methods with baselines on LitBank~(see Table~\ref{tab:Per_com_litbank}), with full results available in Table~\ref{tab:Per_com_litbank_app} of Appendix~\ref{app:add_exp_res}). 
Specifically, in terms of Avg.F1, ImCoref-CeS consistently outperforms all baselines apart from LLMLink, with ImCoref-CeS$_{\text{dpsk}}$ achieving the highest performance. This further validates the superior performance of our method. Notably, although LLMLink attains second-best coreference performance, it requires training multiple LLMs, incurring substantially higher computational costs.
Among supervised neural methods, ImCoref achieves the best Avg. F1 (79.7\%), surpassing Maverick\(_{\text{mes}}\) (78.4\%) and Dual-cache (79.5\%). However, ImCoref ranks second in the MUC metric (87.9\%), slightly behind longdoc and Dual-cache (both 88.2\%). This suggests that while our method excels overall, there remains room for improvement in MUC compared to models specifically designed for long-text coreference. We plan to address this in future work by exploring more suitable mention clustering modules.


\begin{table}[htbp]
  \centering
  \resizebox{1.0\columnwidth}{!}{
    \begin{tabular}{lccc|c}
    \hline
    \multicolumn{1}{c}{Methods} & MUC   & B$^3$   & CEAF$_{\phi_4}$ & Avg.F1 \\
    \hline
    longdoc & -     & -     & -     & 60.1  \\
    Dual-cache & 72.1  & 62.1  & 54.7  & 63.0  \\
    LingMess & -     & -     & -     & 62.6  \\
    Maverick$_{\text{mes}}$ & \textbf{81.5}  & 65.4  & 53.5  & 66.8  \\
    ImCoref(\textbf{ours}) & 80.6  & \textbf{66.7}  & \textbf{55.6}  & \textbf{67.6}  \\
    \hline
    InstructGPT & -     & -     & -     & 72.9  \\
    GPT-3.5-turbo & -     & -     & -     & 70.8  \\
    \hline
    ImCoref-CeS$_{\text{qwen}}$ & 81.8  & 67.3  & 65.0  & 71.3 \\
    ImCoref-CeS$_{\text{dpsk}}$ & 82.4  & 69.2  & 65.5  & 72.4 \\
    ImCoref-CeS$_{\text{gpt4}}$ & \textbf{83.6}  & \textbf{69.8}  & \textbf{66.1}  & \textbf{73.2}  \\
    \hline
    \end{tabular}%
    }
    \caption{Performance comparison (\%) of different methods on WikiCoref. Please see Table~\ref{tab:Per_com_wikiCoref_app} in Appendix~\ref{app:add_exp_res} for detailed results.}
  \label{tab:Per_com_wikiCoref}%
  \vspace*{-1ex}
\end{table}%

\textbf{WikiCoref.}
To assess generalization, we apply models trained on OntoNotes to \textit{\textbf{out-of-domain}} WikiCoref and compare them with existing competitors, as shown in Table~\ref{tab:Per_com_wikiCoref}. The results demonstrate that ImCoref-CeS$_{\text{gpt4}}$ achieves the best performance across all metrics, with an Avg. F1 of 73.2\%. 
Compared to ImCoref, LLM-enhanced ImCoref-CeS variants yield an average improvement of 4.7\% in Avg. F1, indicating that incorporating LLMs as reasoning-augmented components during inference significantly strengthens cross-domain generalization. 
Furthermore, pure LLM-based methods such as InstructGPT and GPT-3.5-turbo also outperform all supervised neural models in Avg. F1, highlighting the generalization limitations of supervised neural methods that are trained on in-domain data. Our proposed framework effectively mitigates this constraint by leveraging the reasoning capacity of LLMs without requiring additional training, thus enabling better adaptation to out-of-domain scenarios.

\subsection{Ablation Study}

\begin{figure}[ht]
  \centering
  \includegraphics[width=1.0\linewidth]{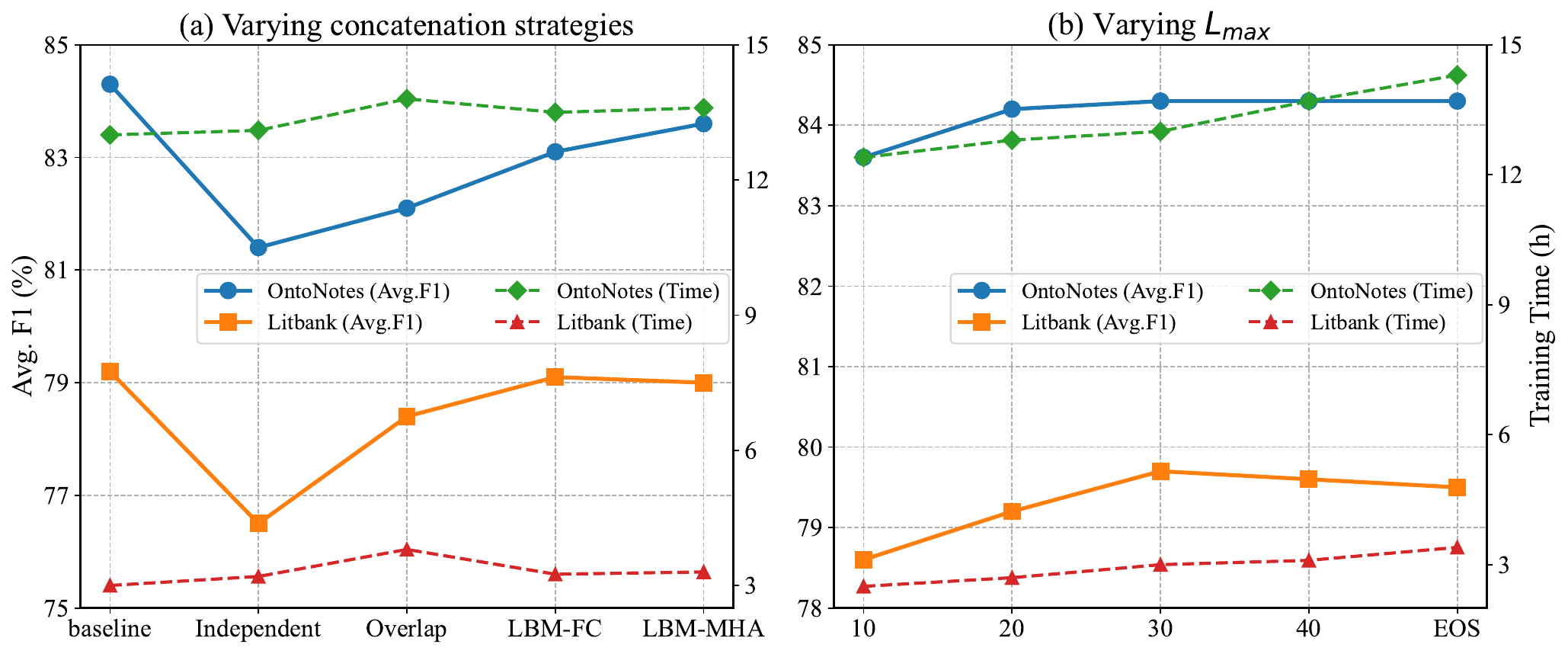}
  \caption{Avg.F1 (\%) and Training Time (h) with varying concatenation strategies and $L_{\text{max}}$. } 
  \label{pic2_fig:}
\end{figure}

\textbf{The Efficacy of LBM and Varying $L_{\text{max}}$.} We look into  the efficacy of LBM in long-text scenarios, and the impact of varying $L_{\text{max}}$ in HyMR by employing ImCoref.
For the \textit{\textbf{former}}, we simulate long-text processing by segmenting documents that exceed the maximum input length ($T = 512$). Although our base encoder DeBERTa$_\text{large}$ can process long texts, it incurs high computational costs when handling extremely long sequences. Therefore, the segmentation strategies remain necessary for efficient long-text processing. 
We compare two LBM variants (i.e., LBM-FC and LBM-MHA) against existing segmentation strategies \textit{independent} and \textit{overlapping} ($T/2$). 
From Fig.~\ref{pic2_fig:}(a), text segmentation inevitably leads to a decline in coreference performance. 
However, both LBM-FC and LBM-MHA consistently outperform the baseline strategies w.r.t. Avg. F1, without substantially increasing training time.
This confirms that LBM enhances encoding capability, thus augmenting ImCoref. 
See more results in Table~\ref{tab:concatenation_strategies_app} of Appendix~\ref{app:add_exp_res}.
For the \textit{\textbf{latter}}, we select $L_{\text{max}}$ from $\{10, 20, 30, 40, +\infty\}$ (Note: when $L_{\text{max}}=+\infty$, $L_s = L_{\text{EOS}s}$). Fig.~\ref{pic2_fig:}(b) reports the Avg.F1 under different $L_{\text{max}}$ settings. 
On OntoNotes, Avg. F1 initially increases with $L_{\text{max}}$ and then stabilizes. On LitBank, it rises initially but slightly declines after reaching a peak. 
Notably, $L_{\text{max}}$ = 30 yields the optimal performance on both datasets. 
Furthermore, training time increases monotonically with larger \(L_{\text{max}}\) values.
These results indicate that HyMR effectively conserves computational resources during training without compromising Avg.F1. See Table~\ref{tab:lmax_app} in Appendix~\ref{app:add_exp_res} for more results.

\textbf{Necessity of LLM Checker-Splitter.} We delve into each component of LLM Checker-Splitter by employing ImCoref-CeS$_{\text{gpt4}}$. During inference, the LLM serves two roles: the Mention Checker (M-Ce), which refines mention detection, and the Coreference Checker-Splitter (C-CeS), which adjusts mention clustering.
We conduct a leave-one-out ablation by removing M-Ce or C-CeS individually, and also evaluate the base model ImCoref, which lacks both components.
As shown in Table~\ref{tab:ceco}, removing either M-Ce or C-CeS leads to a drop in Avg. F1, and the absence of both causes a further decline. This confirms that both components are essential to ImCoref-CeS$_{\text{gpt4}}$.  
Notably, removing C-CeS has a stronger negative impact than removing M-Ce, suggesting that directly correcting coreference clusters contributes more significantly to performance improvement compared to filtering mentions alone.

\begin{table}[htbp]
  \centering
  \resizebox{1.0\columnwidth}{!}{
    \begin{tabular}{lccc|c}
    \hline
    \multirow{2}[2]{*}{} & \multicolumn{3}{c|}{Mention Metric} & \multirow{2}[2]{*}{Avg.F1} \\
          & P     & R     & F1    &  \\
    \hline
    \rowcolor{Gray}\multicolumn{5}{l}{OntoNotes} \\
    ImCoref-CeS$_{\text{gpt4}}$ & 96.3  & 94.6  & 95.4  & \textbf{86.0}  \\
    \hdashline
    -w/o. M-Ce & 94.6  & 94.8  & 94.9  & 85.0  \\
    -w/o. C-CeS & 96.3  & 94.6  & 95.4  & 84.5  \\
    ImCoref & 94.6  & 94.8  & 94.9  & 84.3  \\
    \hline
    \rowcolor{Gray}\multicolumn{5}{l}{LitBank} \\
    ImCoref-CeS$_{\text{gpt4}}$ & 94.3  & 92.1  & 93.2  & \textbf{81.1}  \\
    \hdashline
    -w/o. M-Ce & 92.7  & 92.3  & 92.5  & 80.5  \\
    -w/o. C-CeS & 94.3  & 92.1  & 93.2  & 79.9  \\
    ImCoref & 92.7  & 92.3  & 92.5  & 79.7  \\
    \hline
    \end{tabular}%
    }
    \caption{Impact of each component for LLM Checker-Splitter. See Table~\ref{tab:ceco_app} in Appendix~\ref{app:add_exp_res} for more results.}
  \label{tab:ceco}%
\end{table}%

\begin{figure}[ht]
  \centering
  \includegraphics[width=1.0\linewidth]{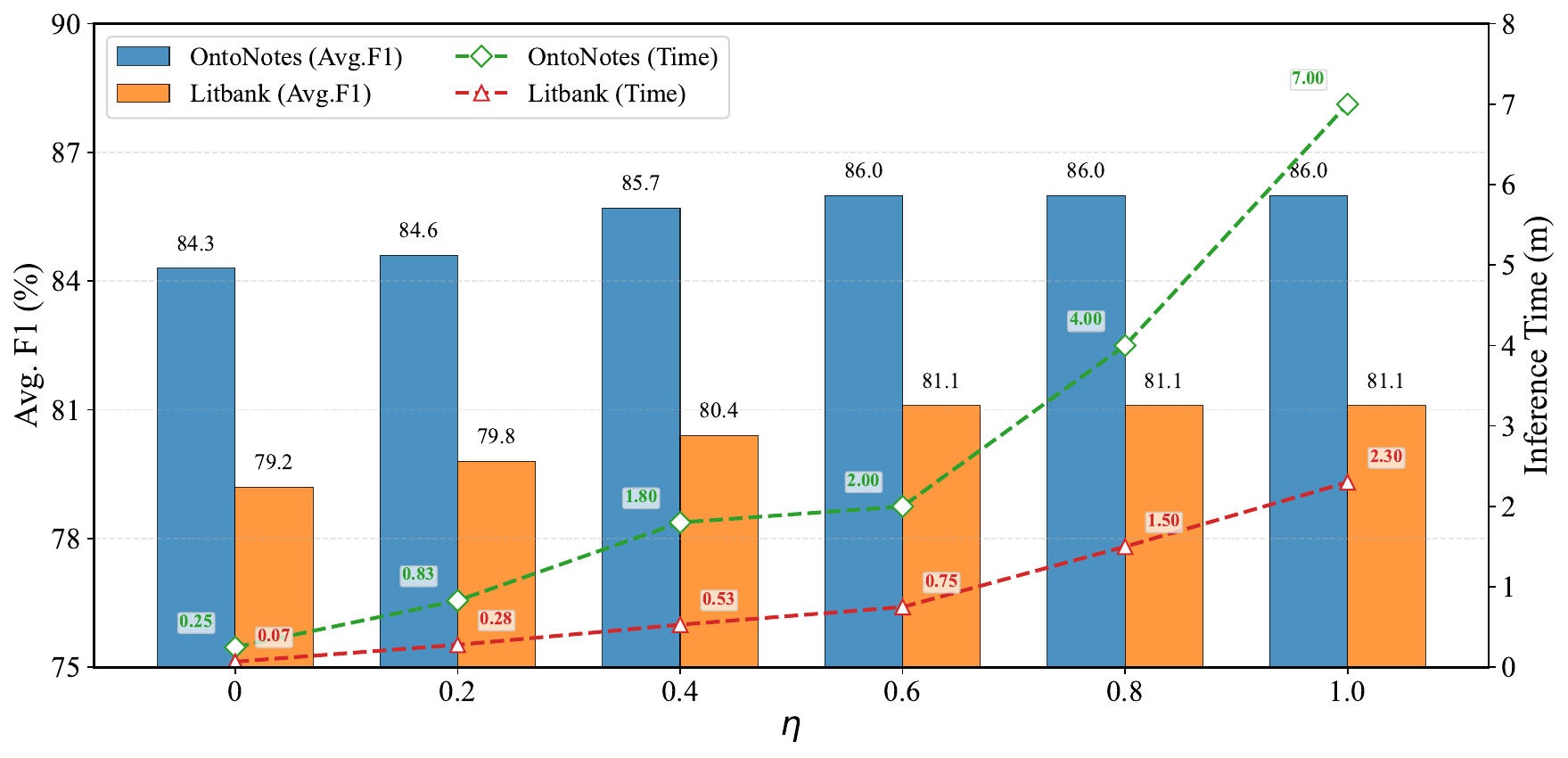}
  \caption{Avg.F1 (\%) and Inference Time (m) with varying $\eta$. } 
  \label{pic3_eta:}
\end{figure}

\textbf{Utility of Mention and Coreference Cluster Filters.} 
We perform an ablation study on OntoNotes and LitBank using ImCoref-CeS$_{\text{gpt4}}$ to assess the influence of $\eta$, which controls the application of the LLM Checker-Splitter. The value of $\eta$ is varied over $\{0, 0.2, 0.4, 0.6, 0.8, 1.0\}$. 
As illustrated in Fig.~\ref{pic3_eta:}, Avg. F1 score initially increases and then plateaus as $\eta$ grows. Specifically, performance on both OntoNotes and LitBank improves steadily when $\eta$ increases from $0$ to $0.6$, but saturates for $\eta > 0.6$. 
In contrast, inference time increases monotonically and substantially with $\eta$.
These results demonstrate that while integrating the LLM Checker-Splitter effectively boosts coreference resolution performance, applying it exhaustively (i.e., $\eta = 1.0$) is computationally inefficient, as it incurs considerable latency without additional gains in Avg. F1. 
The saturation of performance beyond $\eta = 0.6$ underscores the effectiveness of using mention and coreference cluster filters to strike an optimal trade-off between accuracy and efficiency. Hence, selecting an intermediate $\eta$ value is critical for maintaining high performance while minimizing computational overhead. 
\textit{\textbf{Of note, we report the performance of ImCoref-CeS$_{\text{gpt4}}$ across varing $\rho$ (see  Fig.~\ref{pic_rho:app} in Appendix~\ref{app:add_exp_res}).}}

\section{Related Works}
\label{Related_works:app}
\subsection{Coreference Resolution}
CR has long been a crucial task in the field of NLP. The seminal work by~\cite{lee2017end} introduced the first end-to-end supervised neural method, i.e., the Coarse-to-Fine model, establishing the detect-then-cluster pipeline. 
Shortly thereafter, a panoply of efforts have focused on improving this method's training efficiency, memory consumption, and coreference performance. 
For example, some methods~\cite{dobrovolskii-2021-word, kantor2019coreference, xu-choi-2020-revealing, wu-etal-2020-corefqa, kirstain2021coreference,lai2022end, otmazgin2022f,otmazgin2022lingmess,d2023caw,sundar-etal-2024-major} retained the detect-then-cluster pipeline and leveraged pre-trained language models (e.g., BERT~\cite{devlin-etal-2019-bert}, SpanBERT~\cite{joshi2020spanbert}, Longformer~\cite{beltagy2020longformer}) for document encoding, significantly boosting performance. However, they still grapple with substantial memory overhead. Concurrently, there exists another line of works~\cite{xia-etal-2020-incremental, toshniwal-etal-2020-learning, guo-etal-2023-dual, martinelli-etal-2024-maverick} to drew inspiration from human cognitive incremental processing mechanisms. During the mention clustering, they progressively evaluate the association between candidate mentions and existing coreference clusters using linear classifiers or lightweight Transformer architectures to determine coreference links. Such incremental methods require manual tuning of memory configurations to balance coreference performance against training time. 
Notably, generative methods~\cite{liu-etal-2022-autoregressive, bohnet-etal-2023-coreference, zhang-etal-2023-seq2seq}, centered around large sequence-to-sequence architectures, once dominated the pursuit of high coreference performance. Nevertheless, their prohibitively expensive training costs and high inference latency render their deployment infeasible.

Recently, Maverick$_{\text{mes}}$ proposed by~\cite{martinelli-etal-2024-maverick} achieves SOTA coreference performance, which improves the mention detection by employing a ``predict start positions of mentions first, then conditionally predict their end positions'' scheme, and introducing an end-of-sentence (EOS) mention regularization strategy. 
However, it suffers from two pitfalls: underutilization of positional information, and the tendency of EOS mention regularization to inject excessively long mentions during training, leading to the prediction of invalid ones.
Furthermore, existing methods~\cite{joshi2019bert, toshniwal-etal-2020-learning, guo-etal-2023-dual, martinelli-etal-2024-maverick} typically harness strategies like \textit{independent} or \textit{overlapping} to segment the text, encode segments independently, and then sequentially concatenate the representations. Nonetheless, this process is the semantic isolation between different text segments during both training and inference, which hinders the model's ability to capture long-distance coreference relations. 

\subsection{LLMs for Coreference Resolution}
With super-sized training corpora and computational cluster resources, LLMs have demonstrated powerful reasoning capabilities, thus enabling SOTA performance in a wide range of natural language tasks~\cite{achiam2023gpt, liu2024deepseek,qwen3technicalreport, luo2024let, tan2024llm, yu2024automated,si2025goal,wang2025document, si2025teaching,luo2025gltw,si2025aligning,si2024gateau, bai2025sis}. 
However, their performance in CR has yet to surpass mainstream supervised neural methods~\cite{le2023large, gan-etal-2024-assessing}. This limitation stems from the ``large but not precise'' nature of LLMs, which hinders high-accuracy mention detection~\cite{liu2024bridging}. Intriguingly, under ideal conditions where gold mentions are provided, LLMs leverage their strong reasoning capabilities to achieve CR performance that matches or even exceeds that of supervised neural methods~\cite{le2023large}. Recently, other attempts to incorporate LLMs into CR either rely on overly idealistic assumptions~\cite{sundar-etal-2024-major} or entail prohibitive computational costs to train multiple LLMs~\cite{zhu-etal-2025-llmlink}.

Despite the success of supervised neural methods, their small model scale and task-specific nature make them prone to generating a large number of invalid mentions when applied to out-of-domain data~\cite{toshniwal2021generalization, xia-van-durme-2021-moving}. Even within their target domain, invalid mentions and coreference errors remain a significant bottleneck for further performance improvement~\cite{martinelli-etal-2024-maverick}. This presents an opportunity to combine the strengths of supervised neural methods and LLMs. \textbf{To our knowledge, our work is the first exploration of a collaborative mechanism specifically for CR between them.}

\section{Conclusion}
In this paper, we propose a novel framework CR, named \ourmethod. First, we present an improved CR method (\ourmethodfirst) to further explore the performance potential of the supervised neural method under constrained computational resources: introducing a lightweight bridging module to enhance long-text encoding efficiency, tailoring a biaffine scorer during mention detection to comprehensively capture positional information, and invoking a hybrid mention regularization strategy to improve training efficiency.
Building upon this, we then integrate LLMs to dynamically collaborate with \ourmethodfirst~during inference. Specifically, LLMs are prompted to check candidate mentions (filtering out invalid ones) and coreference results (splitting erroneous clusters) predicted by \ourmethodfirst. Extensive experiments demonstrate the effectiveness of \ourmethod, which achieves superior performance compared to existing SOTA methods.  

\section*{Limitations}
We acknowledge that the ImCoref-CeS method, despite its demonstrated performance advantages, has several limitations. The study concentrates on advancing mention detection within the supervised neural method (i.e., Maverick), while the design of a high-performance mention clustering algorithm---essential for real-world applicability---warrants further investigation. Furthermore, beyond the capabilities of the LLM Checker-Splitter, intuitively promising strategies like coreference cluster fusion could be explored. Determining how to best trade off the performance improvements from such strategies against their significant resource demands constitutes a promising yet challenging direction for future work.

\section*{Acknowledgments}
This work is supported by the National Natural Science Foundation of China (No.T2341003), Beijing Municipal Science and Technology Plan Project (Z241100001324025) and a grant from the Guoqiang Institute, Tsinghua University.


\bibliography{custom}

\clearpage
\appendix

\section*{Appendix}
\label{sec:appendix}

\noindent This appendix is organized as follows.  

\begin{itemize}
    \item In Section~\ref{annotation_guid:app}, we elaborate on the annotation guidelines for the LLM-based Checker-Splitter component, detailing the specific text-marking schemes used for mention validation and coreference cluster verification and splitting.
    \item In Section~\ref{app:prompt_temp}, we present the complete set of prompt templates employed by the LLM Checker-Splitter, accompanied by concrete input-output examples that demonstrate its operation across different validation and correction scenarios.
    \item In Section~\ref{app:full_pronouns_list}, we list all the pronouns, and illustrate the text segmentation strategies and LBM.
    \item In Section~\ref{app:com_exp_set}, we provide a comprehensive description of the experimental setup, including the datasets employed, baseline methods compared, evaluation metrics utilized, and detailed implementation configurations.
    \item In Section~\ref{app:add_exp_res}, we present supplementary experimental results, including detailed performance comparisons and the impact of $\rho$.
    \item In Section~\ref{app:error_ana}, we conduct a qualitative error analysis by comparing model outputs against gold annotations, highlighting the effectiveness of our method in reducing mention and coreference errors while identifying remaining challenges.
\end{itemize}

\begin{figure*}[ht]
  \centering
  \includegraphics[width=1.0\linewidth]{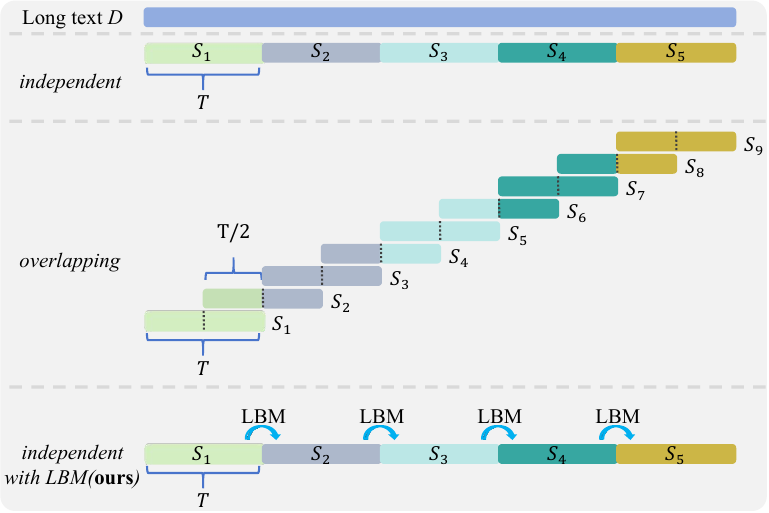}
  \caption{Illustration of text segmentation strategies and LBM: \textit{independent} splits $D$ into non-overlapping segments of length $T$; \textit{overlapping} generates segments via a sliding window with a $T/2$ step size; LBM introduces inter-segment semantic propagation atop \textit{independent}.} 
  \label{pic2_lbm:app}
\end{figure*}

\section{Annotation Guidelines for Checker-Splitter}
\label{annotation_guid:app}
 To clarify, take the input text {\ttfamily\small Jesus called his... God 's kingdom . "} as an example. During mention detection, \ourmethodfirst~outputs candidate mentions formatted as:{\ttfamily\small [...(his,1692,1692),(Jesus,1696,1696),(another man,1699,1700)...(Jesus,1726,1726)...]}. Each candidate includes its text span and start/end positions in the original text. Using these position information, we annotate the mentions in the original text with square brackets ``['' and ``]''. For example, {\ttfamily\small(another man,1699,1700)} is annotated as:{\ttfamily\small "...The birds have nests. But the Son of Man has no place where he can rest his head."Jesus said to [another man],"Follow me!"}. Then, through the specialized prompt template, the LLM act as a mention checker. It checks each candidate's validity based on local context (i.e., the target mention's sentence and several preceding ones) and removes invalid mentions. 
The annotated local text is then concatenated with the prompt template and fed to the LLM for verification, see Fig.~\ref{Framework_fig:}.

During mention clustering, \ourmethodfirst~generates initial coreference clusters from validated mentions, formatted as {\ttfamily\small [...[(his,1692,1692),(Jesus,1696,1696),(Jesus, 1726,1726)...], [(another man,1699,1700) ...]...]}. The LLM then functions as a coreference checker to verify all coreference clusters with two or more mentions. Taking cluster {\ttfamily\small [(his,1692,1692),(Jesus,1696,1696),(Jesus,1726, 1726)...]} as an example, we first number each mention (e.g., {\ttfamily\small \#1:his,\#2:Jesus,\#3:Jesus}). Using their position information, we mark all mentions from the cluster in the original text with ``[(\#X)'' and ``](\#X)'' (X represents the number). 
For instance: {\ttfamily\small "... The birds have nests . But the Son of Man has no place where he can rest [(\#1)his](\#1) head . " [(\#2)Jesus](\#2) said to another man ... But [(\#3)Jesus](\#3) said to him...}. This mark scheme ensures each mention has a unique position, preventing confusion from mentions with the same form. For verification, only sentences containing the cluster's mentions and intervening mention-free sentences are provided as context. The original text fragment, mention number list, and annotated text fragment are combined with a specific prompt template to form the LLM input.

When the coreference checker identifies an erroneous cluster (i.e., mentions within the cluster do not all refer to the same entity), the LLM acts as a coreference splitter. It regroups the mentions to ensure coreferential consistency within new groups and mutual exclusivity between groups. For correcting individual clusters, we employ the same text annotation manner used during cluster checking. The input for the LLM is illustrated in Fig, \ref{Framework_fig:}.


\section{Prompt Templates and Samples of LLM Checker-Splitter}
\label{app:prompt_temp}

This section presents the complete prompt templates for the LLM Checker-Splitter framework along with demonstrative examples. Specifically, Table~\ref{tab:example1_valid_mention_checker} illustrate mention checker implementations, Tables~\ref{tab:example1_cluster_checker}-\ref{tab:example2_cluster_checker} demonstrate coreference checker applications with correct coreference clusters, Tables~\ref{tab:example3_cluster_checker}-\ref{tab:example4_cluster_corrector} provide coreference Checker-Splitter exemplars with incorrect coreference clusters.

\begin{table*}[htbp]
  \centering
  \resizebox{2.0\columnwidth}{!}{
    \begin{tabular}{p{73.91em}}
    \toprule
    Your task is to check whether a mention is valid.\newline{}\newline{}\#\#\# Task Description\newline{}- Input includes: A text segment to be checked, which contains a mention (marked by a pair of ``['' and ``]'').\newline{}- Your task is: To determine whether the span between ``['' and ``]'' accurately represents a valid mention. If yes, output ``Yes''; if not, output ``No''; if uncertain, output    ``Pending''.\newline{}\newline{}\#\#\# Key Definitions\newline{}- \textbf{Mention}: Refers to a specific occurrence in the text of one or more consecutive words used to refer to an entity (person/object/place/organization/concept, etc.), including pronouns, noun phrases, or descriptive spans. A mention must be a sequence of consecutive words, and usually does not include irrelevant punctuation (such as periods, question marks, or commas), unless the punctuation is an integral part of the entity's name (e.g., ``U.S.A.'').\newline{}- \textbf{``['' and ``]'' Markers}: These symbols indicate the boundaries of a mention, where ``['' marks the start position and ``]'' marks the end position, with no space between the markers and adjacent words (e.g., ``[word]'').\newline{}\newline{}\#\#\# Checking Criteria\newline{}Based on the definition of a mention, check the correctness of the marked span:\newline{}1. \textbf{Valid (Output ``Yes'')} if ALL are true:\newline{}   - The span is a complete referring expression of consecutive words.\newline{}   - The span fully captures an entity reference in context.\newline{}   - The span does not include irrelevant punctuation or extra words.\newline{}2. \textbf{Invalid (Output ``No'')} if ANY applies:\newline{}   - The span misses essential modifiers (e.g., articles, quantifiers) needed for entity reference.\newline{}   - The span contains irrelevant punctuation, multiple entities, or non-entity words.\newline{}   - The span is a partial fragment of an entity that cannot independently refer.\newline{}   - The span has boundary errors (incorrect start or end position), making it incomplete or overextended.\newline{}3. \textbf{Uncertain (Output ``Pending'')}: The context is ambiguous and it is not possible to determine whether the span is a valid entity mention.\newline{}\newline{}\#\#\# Output Format Requirements\newline{}- First, output the reason for the judgment (a brief analysis within 20 words).\newline{}- Then, output the judgment result, which must be: ``Yes'', ``No'', or ``Pending''.\newline{}- The output should end with the judgment result, without generating any additional content or explanation.\newline{}- Example output structure: ``Reason text. Yes'',  ``Reason text. No'' or ``Reason text. Pending''. \newline{} \\
    \rowcolor{Gray}
    \#\#\# Input to be Checked\newline{}
    Someone said to Jesus , `` I will follow you anywhere you go .'' He answered , `` The foxes have holes to live in . But the Son of Man has no place where he can rest his head . '' Jesus said to another man , `` Follow me ! '' But the man said , `` Lord , let me go and bury my father first . '' But [Jesus] said to him, `` Let the people who are dead bury their own dead .\\
    LLM Output:\newline{}
    The mention [Jesus] is a single-word proper noun with correct bracket placement and no extraneous punctuation. Yes
    \\
    \midrule
    \rowcolor{Gray}
    \#\#\# Input to be Checked\newline{}
    Nickelodeon 's latest cartoon series is about a boy with a disability . Gloria Hillard takes a look at what the show `` Pelswick '' means to some children . Imagine being able to visit Nickelodeon to see how they make cartoons . Here 's the Beavers and all his friends . Cool . That 's awesome . 15 - year - old Nicky Barischoff and [12 - year - old Allison Gray] are best friends .\\
    LLM Output:\newline{}
    The span includes hyphens as part of "12-year-old" and correctly captures the full referential expression. Yes\\
    \midrule
    \rowcolor{Gray}
    \#\#\# Input to be Checked\newline{}
    The fifth angel blew his trumpet . Then I saw a star fall from the sky to the earth . The star was given the key to [the deep hole that leads down to the bottomless pit] .\\
    LLM Output:\newline{}
    The mention is a complete noun phrase with essential modifiers and correct boundaries. Yes\\
    \midrule
    \rowcolor{Gray}
    \#\#\# Input to be Checked\newline{}
    The Box was covered with gold . Inside this Box was a golden jar of manna and Aaron 's rod -- the rod that once grew leaves . Also in the Box were the flat stones with [the Ten Commandments of the old agreement on them] .\\
    LLM Output:\newline{}
    The span includes extra non-entity words ("on them") and is a dependent fragment. No\\
    \midrule
    \rowcolor{Gray}
    \#\#\# Input to be Checked\newline{}
    And people did to him all the bad things they wanted to do . The Scriptures said this would happen to him . '' Then Jesus , [Peter , James , and John] went to the other followers .\\
    LLM Output:\newline{}
    The span is a partial fragment of the compound subject entity "Jesus, Peter, James, and John" and includes irrelevant structural commas. No.\\
    \midrule
    \rowcolor{Gray}
    \#\#\# Input to be Checked\newline{}
    [They were to hurt only those who did not have God 's mark on their foreheads . They] were not given the power to kill them but only to cause them pain for five months -- pain like a person feels when stung by a scorpion .\\
    LLM Output:\newline{}
    Span captures multiple clauses and includes redundant subject repetition. No.\\
    \bottomrule
    \end{tabular}%
    }
  \caption{Examples of LLM mention checker.}
  \label{tab:example1_valid_mention_checker}
\end{table*}%

\begin{table*}[htbp]
  \centering
  \resizebox{2.0\columnwidth}{!}{
    \begin{tabular}{p{75.5em}}
    \toprule
    Your task is to check whether all mentions in a coreference cluster point to the same entity.\newline{}\newline{}\#\#\# Task Description\newline{}- Input includes: A original text, a coreference cluster result (with a set of numbered mentions), and text with these numbered mentions marked.\newline{}- Your task is: To determine if all mentions in the cluster clearly point to the same entity. If all mentions point to the same entity, output ``Yes''; if at least one mention points to a different entity, output ``No''; if there is insufficient information to determine, output ``Pending''.\newline{}\newline{}\#\#\# Key Definitions\newline{}- \textbf{Mention}: Refers to a specific occurrence in the text of one or more consecutive words used to refer to an entity (person/object/place/organization/concept, etc.), including pronouns, noun phrases, or descriptive spans. A mention must be a sequence of consecutive words, and usually does not include irrelevant punctuation (such as periods, question marks, or commas), unless the punctuation is an integral part of the entity's name (e.g., ``U.S.A.'').\newline{}- \textbf{Coreference Task}: Group mentions at different positions in the text that point to the same entity into one cluster.\newline{}- \textbf{Mention Numbering Format}: Each mention is assigned a unique number in the format ``\#X'', where X is a numeral (e.g., ``\#1: you'' indicates that the mention ``you'' is numbered \#1).\newline{}- \textbf{Mention Annotation Format in Text}: In text with numbered mentions, the start of each mention is marked with ``[(\#X)'' and the end with ``](\#X)'' (where X is a numeral), with no space between the markers and adjacent words (e.g., ``[(\#1)you](\#1)'' represents the mention ``you'' numbered \#1). This ensures the location of each mention instance is unique, avoiding confusion between mentions with the same form.\newline{}\newline{}\#\#\# Checking Criteria\newline{}- \textbf{Determine ``Yes''}: When all mentions clearly refer to the same entity in context.\newline{}- \textbf{Determine ``No''}: When at least one mention in the cluster clearly refers to a different entity.\newline{}- \textbf{Determine ``Pending''}: When the context does not allow confirmation of coreference relationships for all mentions.\newline{}\newline{}\#\#\# Output Format Requirements\newline{}- First, output the reason for the judgment (a brief analysis within 50 words).\newline{}- Then, output the judgment result, which must be: ``Yes'', ``No'', or ``Pending''.\newline{}- The output should end with the judgment result, with no extra content or explanation.\newline{}- Example output structure: ``Reason text. Yes'', ``Reason text. No'', or ``Reason text. Pending'' \\
    \\
    \rowcolor{Gray}
    \#\#\# Input to be Checked\newline{}
    Original text: \newline{}
    As we said , earlier this week , Christine Todd Whitman stepped down and now a whole lot of analysis is taking place as to why she stepped down , the job she 's done when she was in office and the job that has yet to be complete . From our New York bureau as I was introducing a moment ago , Dan Fagen , the environment writer for News Day . and Dan , Whitman did consider herself sort of the Republican environmentalist , but now there 's been an awful lot of Monday night quarterback quarterbacking taking place now that she 's stepped down . What do you suppose took place here ? I think many people have watched this administration in action and watched Christy Whitman in action . The mystery is that she took the job in the first place or did n't leave earlier . She was quite a misfit for this administration , and she showed that over and over . She was not good at all at communicating the administration 's message to the public and certainly to the environmental groups and she was even worse at bringing the EPA bureaucracy , the EPA environmental policy makers ' message to this administration because she saw herself undercut over and over . That 's interesting because leading into it , while she was New Jersey governor , she sued the EPA and she was very much considering herself to be an environmentalist , and then only to end up becoming the head of the EPA and then of course , there was a lot of criticism over relaxed regulations . Let 's hear what she had to say last Wednesday here on CNN just after announcing her resignation . I 'm not leaving because of clashes with the administration , in fact , I have n't had any . I report to the President , and he has always asked me to give him my best unadulterated advice . We have have been on the same page . I was never supportive of protocol because it did n't do what everyone needs to get done as far as climate change is concerned . I 'm leaving now because it 's the appropriate time to do it .
    \newline{}\newline{}
    Coreference cluster result (with a set of numbered mentions): \newline{}
    \#1:stepped, \#2:stepped, \#3:her resignation, \#4:it, \#5:leaving
    \newline{}\newline{}
    Text with these numbered mentions: \newline{}
    As we said , earlier this week , Christine Todd Whitman [(\#1)stepped](\#1) down and now a whole lot of analysis is taking place as to why she [(\#2)stepped](\#2) down , the job she 's done when she was in office and the job that has yet to be complete . From our New York bureau as I was introducing a moment ago , Dan Fagen , the environment writer for News Day . and Dan , Whitman did consider herself sort of the Republican environmentalist , but now there 's been an awful lot of Monday night quarterback quarterbacking taking place now that she 's stepped down . What do you suppose took place here ? I think many people have watched this administration in action and watched Christy Whitman in action . The mystery is that she took the job in the first place or did n't leave earlier . She was quite a misfit for this administration , and she showed that over and over . She was not good at all at communicating the administration 's message to the public and certainly to the environmental groups and she was even worse at bringing the EPA bureaucracy , the EPA environmental policy makers ' message to this administration because she saw herself undercut over and over . That 's interesting because leading into it , while she was New Jersey governor , she sued the EPA and she was very much considering herself to be an environmentalist , and then only to end up becoming the head of the EPA and then of course , there was a lot of criticism over relaxed regulations . Let 's hear what she had to say last Wednesday here on CNN just after announcing [(\#3)her resignation](\#3) . I 'm not leaving because of clashes with the administration , in fact , I have n't had any . I report to the President , and he has always asked me to give him my best unadulterated advice . We have have been on the same page . I was never supportive of protocol because it did n't do what everyone needs to get done as far as climate change is concerned . I 'm [(\#5)leaving](\#5) now because it 's the appropriate time to do [(\#4)it](\#4) .\\
    LLM Output:\newline{}
    The cluster groups ``stepped'' (mentions \#1, \#2), ``her resignation'' (\#3), ``leaving'' (\#5), and ``it'' (\#4). Mentions \#1 and \#2 both refer to Christine Todd Whitman stepping down. Mention \#3 explicitly refers to her resignation. Mention \#5 refers to her act of leaving. Mention \#4 (``it'') refers back to the action of leaving. All mentions consistently point to Whitman's resignation/leaving action. Yes\\
    \bottomrule
    \end{tabular}%
    }
    \caption{Example 1 of LLM coreference Checker-Splitter with the right coreference cluster.}
  \label{tab:example1_cluster_checker}%
\end{table*}%

\begin{table*}[htbp]
  \centering
  \resizebox{2.0\columnwidth}{!}{
    \begin{tabular}{p{75.5em}}
    \toprule
    Your task is to check whether all mentions in a coreference cluster point to the same entity.\newline{}\newline{}\#\#\# Task Description\newline{}- Input includes: A original text, a coreference cluster result (with a set of numbered mentions), and text with these numbered mentions marked.\newline{}- Your task is: To determine if all mentions in the cluster clearly point to the same entity. If all mentions point to the same entity, output ``Yes''; if at least one mention points to a different entity, output ``No''; if there is insufficient information to determine, output ``Pending''.\newline{}\newline{}\#\#\# Key Definitions\newline{}- \textbf{Mention}: Refers to a specific occurrence in the text of one or more consecutive words used to refer to an entity (person/object/place/organization/concept, etc.), including pronouns, noun phrases, or descriptive spans. A mention must be a sequence of consecutive words, and usually does not include irrelevant punctuation (such as periods, question marks, or commas), unless the punctuation is an integral part of the entity's name (e.g., ``U.S.A.'').\newline{}- \textbf{Coreference Task}: Group mentions at different positions in the text that point to the same entity into one cluster.\newline{}- \textbf{Mention Numbering Format}: Each mention is assigned a unique number in the format ``\#X'', where X is a numeral (e.g., ``\#1: you'' indicates that the mention ``you'' is numbered \#1).\newline{}- \textbf{Mention Annotation Format in Text}: In text with numbered mentions, the start of each mention is marked with ``[(\#X)'' and the end with ``](\#X)'' (where X is a numeral), with no space between the markers and adjacent words (e.g., ``[(\#1)you](\#1)'' represents the mention ``you'' numbered \#1). This ensures the location of each mention instance is unique, avoiding confusion between mentions with the same form.\newline{}\newline{}\#\#\# Checking Criteria\newline{}- \textbf{Determine ``Yes''}: When all mentions clearly refer to the same entity in context.\newline{}- \textbf{Determine ``No''}: When at least one mention in the cluster clearly refers to a different entity.\newline{}- \textbf{Determine ``Pending''}: When the context does not allow confirmation of coreference relationships for all mentions.\newline{}\newline{}\#\#\# Output Format Requirements\newline{}- First, output the reason for the judgment (a brief analysis within 50 words).\newline{}- Then, output the judgment result, which must be: ``Yes'', ``No'', or ``Pending''.\newline{}- The output should end with the judgment result, with no extra content or explanation.\newline{}- Example output structure: ``Reason text. Yes'', ``Reason text. No'', or ``Reason text. Pending'' \\
    \\
    \rowcolor{Gray}
    \#\#\# Input to be Checked\newline{}
    Original text: \newline{}
This role reversal holds true , as well , for his three liberal and moderate allies , Justices Thurgood Marshall , Harry Blackmun and John Stevens . But are these four players , three of them in their 80s , ready to assume a different role after 88 years , collectively , of service on the high court ? Every indication is that the four are prepared to accept this new role , and the frustrations that go with it , but in different ways . Justices Brennan and Stevens appear philosophical about it ; Justices Marshall and Blackmun appear fighting mad . The four justices are no newcomers to dissent , often joining forces in the past decade to criticize the court 's conservative drift . But always , in years past , they have bucked the trend and have been able to pick up a fifth vote to eke out a number of major victories in civil rights and liberties cases . Now , however , as the court 's new five - member conservative majority continues to solidify , victories for the liberals are rare . The change is most dramatic for Justice Brennan , the last survivor of the mid-1960s liberal majority under Chief Justice Earl Warren . In the seven Supreme Court terms from the fall of 1962 through the spring of 1967 , the height of the Warren Court 's power , Justice Brennan cast only 25 dissenting votes in 555 cases decided by the court . Last term alone he cast 52 dissenting votes in 133 decisions , with the contentious flag - burning ruling as his only big victory . But Justice Brennan foresaw his new role , strongly defending the importance of dissents in a 1985 speech . `` Each time the court revisits an issue , the justices will be forced by a dissent to reconsider the fundamental questions and to rethink the result , '' he said . Moreover , in recent months he has said that when he was on the winning side in the 1960s , he knew that the tables might turn in the future . He has said that he now knows how Justice John Harlan felt , a reference to the late conservative justice who was the most frequent dissenter from the Warren Court 's opinions . Associates of 81 - year - old Justice Marshall say he was `` depressed '' about the court 's direction last spring , but is feisty about his role and determined to speak out against the court 's cutbacks in civil rights . `` We could sweep it under the rug and hide it , but I 'm not going to do it , '' he said in a speech last month . He , like Justice Brennan , considers dissents highly important for the future , a point that has n't escaped legal scholars . Harvard Law School Professor Laurence Tribe says there is a `` generation - skipping '' flavor to current dissents . The dissenters in the Warren Court , he says , appeared to be writing for the short - term , suggesting that the court 's direction might change soon . `` Brennan and Marshall are speaking in their dissents to a more distant future , '' he says . Justice Blackmun , who will turn 81 next month , also seems feisty about his new role . Associates say he takes some defeats more personally than his colleagues , especially attempts to curtail the right to abortion first recognized in his 1973 opinion , Roe vs. Wade . Friends and associates who saw Justice Blackmun during the summer said he was no more discouraged about the court than in recent years . And his outlook improved after successful cataract surgery in August . But his level of frustration showed in a recent , impassioned speech to a group of hundreds of lawyers in Chicago . He concluded his remarks by quoting , emotionally and at some length , according to those present , the late Martin Luther King 's famous `` I Have a Dream '' speech from the 1963 March on Washington .
\newline{}\newline{}
Coreference cluster result (with a set of numbered mentions): \newline{}
\#1:Harry Blackmun, \#2:Blackmun, \#3:his, \#4:Justice Blackmun , who will turn 81 next month ,, \#5:his, \#6:he, \#7:his, \#8:he, \#9:Justice Blackmun, \#10:his, \#11:his, \#12:He, \#13:his
\newline{}\newline{}
Text with these numbered mentions: 
This role reversal holds true , as well , for his three liberal and moderate allies , Justices Thurgood Marshall , [(\#1)Harry Blackmun](\#1) and John Stevens . But are these four players , three of them in their 80s , ready to assume a different role after 88 years , collectively , of service on the high court ? Every indication is that the four are prepared to accept this new role , and the frustrations that go with it , but in different ways . Justices Brennan and Stevens appear philosophical about it ; Justices Marshall and [(\#2)Blackmun](\#2) appear fighting mad . The four justices are no newcomers to dissent , often joining forces in the past decade to criticize the court 's conservative drift . But always , in years past , they have bucked the trend and have been able to pick up a fifth vote to eke out a number of major victories in civil rights and liberties cases . Now , however , as the court 's new five - member conservative majority continues to solidify , victories for the liberals are rare . The change is most dramatic for Justice Brennan , the last survivor of the mid-1960s liberal majority under Chief Justice Earl Warren . In the seven Supreme Court terms from the fall of 1962 through the spring of 1967 , the height of the Warren Court 's power , Justice Brennan cast only 25 dissenting votes in 555 cases decided by the court . Last term alone he cast 52 dissenting votes in 133 decisions , with the contentious flag - burning ruling as his only big victory . But Justice Brennan foresaw his new role , strongly defending the importance of dissents in a 1985 speech . `` Each time the court revisits an issue , the justices will be forced by a dissent to reconsider the fundamental questions and to rethink the result , '' he said . Moreover , in recent months he has said that when he was on the winning side in the 1960s , he knew that the tables might turn in the future . He has said that he now knows how Justice John Harlan felt , a reference to the late conservative justice who was the most frequent dissenter from the Warren Court 's opinions . Associates of 81 - year - old Justice Marshall say he was `` depressed '' about the court 's direction last spring , but is feisty about his role and determined to speak out against the court 's cutbacks in civil rights . `` We could sweep it under the rug and hide it , but I 'm not going to do it , '' he said in a speech last month . He , like Justice Brennan , considers dissents highly important for the future , a point that has n't escaped legal scholars . Harvard Law School Professor Laurence Tribe says there is a `` generation - skipping '' flavor to current dissents . The dissenters in the Warren Court , he says , appeared to be writing for the short - term , suggesting that the court 's direction might change soon . `` Brennan and Marshall are speaking in their dissents to a more distant future , '' he says . [(\#4)Justice Blackmun , who will turn 81 next month ,](\#4) also seems feisty about [(\#3)his](\#3) new role . Associates say [(\#6)he](\#6) takes some defeats more personally than [(\#7)his](\#7) colleagues , especially attempts to curtail the right to abortion first recognized in [(\#5)his](\#5) 1973 opinion , Roe vs. Wade . Friends and associates who saw [(\#9)Justice Blackmun](\#9) during the summer said [(\#8)he](\#8) was no more discouraged about the court than in recent years . And [(\#10)his](\#10) outlook improved after successful cataract surgery in August . But [(\#11)his](\#11) level of frustration showed in a recent , impassioned speech to a group of hundreds of lawyers in Chicago . [(\#12)He](\#12) concluded [(\#13)his](\#13) remarks by quoting , emotionally and at some length , according to those present , the late Martin Luther King 's famous `` I Have a Dream '' speech from the 1963 March on Washington .
   \\
    LLM Output:\newline{}
    All mentions in the cluster (\#1: Harry Blackmun, \#2: Blackmun, \#3: his, \#4: Justice Blackmun..., \#5-\#13: his/he/He) explicitly refer to Justice Harry Blackmun. The descriptive mention \#4 (age detail) and subsequent pronouns consistently align with his identity, judicial role, opinions (e.g., Roe vs. Wade), and personal experiences (e.g., surgery, speeches). No mention points to a different entity. Yes.
    \\
    \bottomrule
    \end{tabular}%
    }
    \caption{Example 2 of LLM coreference Checker-Splitter with the right coreference cluster.}
  \label{tab:example2_cluster_checker}%
\end{table*}%

\begin{table*}[htbp]
  \centering
  \resizebox{2.0\columnwidth}{!}{
    \begin{tabular}{p{75.5em}}
    \toprule
    Your task is to check whether all mentions in a coreference cluster point to the same entity.\newline{}\newline{}\#\#\# Task Description\newline{}- Input includes: A original text, a coreference cluster result (with a set of numbered mentions), and text with these numbered mentions marked.\newline{}- Your task is: To determine if all mentions in the cluster clearly point to the same entity. If all mentions point to the same entity, output ``Yes''; if at least one mention points to a different entity, output ``No''; if there is insufficient information to determine, output ``Pending''.\newline{}\newline{}\#\#\# Key Definitions\newline{}- \textbf{Mention}: Refers to a specific occurrence in the text of one or more consecutive words used to refer to an entity (person/object/place/organization/concept, etc.), including pronouns, noun phrases, or descriptive spans. A mention must be a sequence of consecutive words, and usually does not include irrelevant punctuation (such as periods, question marks, or commas), unless the punctuation is an integral part of the entity's name (e.g., ``U.S.A.'').\newline{}- \textbf{Coreference Task}: Group mentions at different positions in the text that point to the same entity into one cluster.\newline{}- \textbf{Mention Numbering Format}: Each mention is assigned a unique number in the format ``\#X'', where X is a numeral (e.g., ``\#1: you'' indicates that the mention ``you'' is numbered \#1).\newline{}- \textbf{Mention Annotation Format in Text}: In text with numbered mentions, the start of each mention is marked with ``[(\#X)'' and the end with ``](\#X)'' (where X is a numeral), with no space between the markers and adjacent words (e.g., ``[(\#1)you](\#1)'' represents the mention ``you'' numbered \#1). This ensures the location of each mention instance is unique, avoiding confusion between mentions with the same form.\newline{}\newline{}\#\#\# Checking Criteria\newline{}- \textbf{Determine ``Yes''}: When all mentions clearly refer to the same entity in context.\newline{}- \textbf{Determine ``No''}: When at least one mention in the cluster clearly refers to a different entity.\newline{}- \textbf{Determine ``Pending''}: When the context does not allow confirmation of coreference relationships for all mentions.\newline{}\newline{}\#\#\# Output Format Requirements\newline{}- First, output the reason for the judgment (a brief analysis within 50 words).\newline{}- Then, output the judgment result, which must be: ``Yes'', ``No'', or ``Pending''.\newline{}- The output should end with the judgment result, with no extra content or explanation.\newline{}- Example output structure: ``Reason text. Yes'', ``Reason text. No'', or ``Reason text. Pending'' \\
    \\
    \rowcolor{Gray}
    \#\#\# Input to be Checked\newline{}
    Original text: \newline{}
The government sold the deposits of four savings - and - loan institutions , in its first wave of sales of big , sick thrifts , but low bids prevented the sale of a fifth . The four S\&Ls were sold to large banks , as was the case with most of the 28 previous transactions initiated by the Resolution Trust Corp. since it was created in the S\&L bailout legislation two months ago . Two of the four big thrifts were sold to NCNB Corp. , Charlotte , N.C. , which has aggressively expanded its markets , particularly in Texas and Florida . A Canadian bank bought another thrift , in the first RTC transaction with a foreign bank . Under these deals , the RTC sells just the deposits and the healthy assets . These `` clean - bank '' transactions leave the bulk of bad assets , mostly real estate , with the government , to be sold later . In these four , for instance , the RTC is stuck with \$ 4.51 billion in bad assets . Acquirers paid premiums ranging from 1.5 \% to 3.7 \% for the deposits and branch systems , roughly in line with what analysts were expecting . The buyers will also be locked into deposit rates for just two weeks , as has been the case with previous deals . After that , the buyers may repudiate the rates paid by the former thrifts .
\newline{}\newline{}
Coreference cluster result (with a set of numbered mentions): \newline{}
\#1:The government, \#2:four savings - and - loan institutions, \#3:its, \#4:The four S\&Ls, \#5:the four big thrifts, \#6:the government, \#7:these four, \#8:the former thrifts
\newline{}\newline{}
Text with these numbered mentions: \newline{}
[(\#1)The government](\#1) sold the deposits of [(\#2)four savings - and - loan institutions](\#2) , in [(\#3)its](\#3) first wave of sales of big , sick thrifts , but low bids prevented the sale of a fifth . [(\#4)The four S\&Ls](\#4) were sold to large banks , as was the case with most of the 28 previous transactions initiated by the Resolution Trust Corp. since it was created in the S\&L bailout legislation two months ago . Two of [(\#5)the four big thrifts](\#5) were sold to NCNB Corp. , Charlotte , N.C. , which has aggressively expanded its markets , particularly in Texas and Florida . A Canadian bank bought another thrift , in the first RTC transaction with a foreign bank . Under these deals , the RTC sells just the deposits and the healthy assets . These `` clean - bank '' transactions leave the bulk of bad assets , mostly real estate , with [(\#6)the government](\#6) , to be sold later . In [(\#7)these four](\#7) , for instance , the RTC is stuck with \$ 4.51 billion in bad assets . The buyers will also be locked into deposit rates for just two weeks , as has been the case with previous deals . After that , the buyers may repudiate the rates paid by [(\#8)the former thrifts](\#8) .
   \\
    LLM Output:\newline{}
    The cluster combines mentions of the government (e.g., \#1, \#3, \#6) and the four S\&Ls (e.g., \#2, \#4, \#5, \#7, \#8). Pronouns like ``its'' (\#3) and ``the four big thrifts'' (\#5) clearly refer to the government and the S\&Ls respectively. These are distinct entities. No.
    \\
    \bottomrule
    \end{tabular}%
    }
    \caption{Example 3 of LLM coreference Checker-Splitter with the incorrect coreference cluster (\textbf{coreference checker}).}
  \label{tab:example3_cluster_checker}%
\end{table*}%

\begin{table*}[htbp]
  \centering
  \resizebox{2.0\columnwidth}{!}{
    \begin{tabular}{p{88.065em}}
    \toprule
    Your task is to regroup a coreference cluster.\newline{}\newline{}\#\#\# Task Description\newline{}- Input includes: A original text, a coreference cluster result (with a set of numbered mentions), and text with these numbered mentions marked.\newline{}- Your task is: The input coreference cluster has been identified as incorrect (not all mentions point to the same entity). You need to regroup these mentions based on the text context, ensuring each new group's mentions refer to the same entity, and different groups' mentions refer to different entities. The output must include the reasons for regrouping and the regrouping results.\newline{}\newline{}\#\#\# Key Definitions\newline{}- \textbf{Mention}: Refers to a specific occurrence in the text of one or more consecutive words used to refer to an entity (person/object/place/organization/concept, etc.), including pronouns, noun phrases, or descriptive spans. A mention must be a sequence of consecutive words, and usually does not include irrelevant punctuation (such as periods, question marks, or commas), unless the punctuation is an integral part of the entity's name (e.g., ``U.S.A.'').\newline{}- \textbf{Coreference Task}: Group mentions at different positions in the text that point to the same entity into one cluster.\newline{}- \textbf{Mention Numbering Format}: Each mention is assigned a unique number in the format``\#X'', where X is a numeral (e.g., ``\#1: you'' indicates that the mention ``you'' is numbered \#1).\newline{}- \textbf{Mention Annotation Format in Text}: In text with numbered mentions, the start of each mention is marked with ``[(\#X)'' and the end with ``](\#X)'' (where X is a numeral), with no space between the markers and adjacent words (e.g., ``[(\#1)you](\#1)'' represents the mention ``you'' numbered \#1).  This ensures the location of each mention instance is unique, avoiding confusion between mentions with the same form.\newline{}- \textbf{Mention Annotation Method}: All mentions in the input are marked in the above format. Only these numbered mentions can be used for grouping. Unmarked parts of the text only provide context.\newline{}\newline{}\#\#\# Regrouping Criteria\newline{}Strictly follow these rules for regrouping:\newline{}- \textbf{Intra-group Consistency}: All mentions in a new group must clearly refer to the same entity.\newline{}- \textbf{Inter-group Exclusivity}: Mentions in different groups must refer to different entities.\newline{}- \textbf{Completeness}: The regrouping must include all mentions from the original cluster without omission.\newline{}- \textbf{Conflict Resolution}: If a mention's affiliation is unclear, place it in a new group.\newline{}- \textbf{Scope Limitation}: Only process mentions in the input cluster; do not introduce other unmarked mentions from the text.\newline{}- \textbf{Nested Mention Handling}: For nested mentions (one mention containing another), determine their affiliation based on context: group them together if they refer to the same entity, otherwise separate them.\newline{}- \textbf{Failure Handling}: If correction is impossible (e.g., insufficient context or ambiguous mention reference), return the reason for failure. Otherwise, output the reasons and results.\newline{}\newline{}\#\#\# Output Format Requirements\newline{}- First, provide a brief analysis (within 100 words) of the original cluster's error and the regrouping basis.\newline{}- Then, output the regrouped mention numbers. The format is comma-separated groups, each group expressed as: [\#list of numbers] (e.g., [\#1,\#2,\#4], [\#3,\#5]). Numbers must be in ascending order and only from the original cluster.\newline{}- The output must end with the grouping result, with no extra content or explanation.\newline{}- Example output structure: ``Reason text. [\#1, ...], [\#X, ..], ...''.\newline{}- If correction fails, output ``Correction failed:'' followed by the reason (e.g., ``Correction failed: Insufficient context to determine the entity reference of \#3''), then end the output. \\
     \\
    \rowcolor{Gray}
    \#\#\# Input to be Regrouped\newline{}
    Original text: \newline{}
The government sold the deposits of four savings - and - loan institutions , in its first wave of sales of big , sick thrifts , but low bids prevented the sale of a fifth . The four S\&Ls were sold to large banks , as was the case with most of the 28 previous transactions initiated by the Resolution Trust Corp. since it was created in the S\&L bailout legislation two months ago . Two of the four big thrifts were sold to NCNB Corp. , Charlotte , N.C. , which has aggressively expanded its markets , particularly in Texas and Florida . A Canadian bank bought another thrift , in the first RTC transaction with a foreign bank . Under these deals , the RTC sells just the deposits and the healthy assets . These `` clean - bank '' transactions leave the bulk of bad assets , mostly real estate , with the government , to be sold later . In these four , for instance , the RTC is stuck with \$ 4.51 billion in bad assets . Acquirers paid premiums ranging from 1.5 \% to 3.7 \% for the deposits and branch systems , roughly in line with what analysts were expecting . The buyers will also be locked into deposit rates for just two weeks , as has been the case with previous deals . After that , the buyers may repudiate the rates paid by the former thrifts .
\newline{}\newline{}
Coreference cluster result (with a set of numbered mentions): \newline{}
\#1:The government, \#2:four savings - and - loan institutions, \#3:its, \#4:The four S\&Ls, \#5:the four big thrifts, \#6:the government, \#7:these four, \#8:the former thrifts
\newline{}\newline{}
Text with these numbered mentions: \newline{}
[(\#1)The government](\#1) sold the deposits of [(\#2)four savings - and - loan institutions](\#2) , in [(\#3)its](\#3) first wave of sales of big , sick thrifts , but low bids prevented the sale of a fifth . [(\#4)The four S\&Ls](\#4) were sold to large banks , as was the case with most of the 28 previous transactions initiated by the Resolution Trust Corp. since it was created in the S\&L bailout legislation two months ago . Two of [(\#5)the four big thrifts](\#5) were sold to NCNB Corp. , Charlotte , N.C. , which has aggressively expanded its markets , particularly in Texas and Florida . A Canadian bank bought another thrift , in the first RTC transaction with a foreign bank . Under these deals , the RTC sells just the deposits and the healthy assets . These `` clean - bank '' transactions leave the bulk of bad assets , mostly real estate , with [(\#6)the government](\#6) , to be sold later . In [(\#7)these four](\#7) , for instance , the RTC is stuck with \$ 4.51 billion in bad assets . The buyers will also be locked into deposit rates for just two weeks , as has been the case with previous deals . After that , the buyers may repudiate the rates paid by [(\#8)the former thrifts](\#8) .
   \\
    LLM Output:\newline{}
    The original cluster incorrectly grouped government mentions with thrifts/S\&Ls. ``The government'' (\#1,\#3,\#6) refers to the governing entity managing the RTC, while ``four savings-and-loan institutions'' (\#2), ``The four S\&Ls'' (\#4), ``the four big thrifts'' (\#5),   ``these four'' (\#7), and ``the former thrifts'' (\#8) all reference the same four financial institutions involved in the sales. These entities are distinct and must be separated. [\#1,\#3,\#6], [\#2,\#4,\#5,\#7,\#8]
    \\
    \bottomrule
    \end{tabular}%
    }
    \caption{Example 3 of LLM coreference Checker-Splitter with the incorrect coreference cluster (\textbf{coreference corrector}).}
  \label{tab:example3_cluster_corrector}%
\end{table*}%

\begin{table*}[htbp]
  \centering
  \resizebox{2.0\columnwidth}{!}{
    \begin{tabular}{p{75.5em}}
    \toprule
    Your task is to check whether all mentions in a coreference cluster point to the same entity.\newline{}\newline{}\#\#\# Task Description\newline{}- Input includes: A original text, a coreference cluster result (with a set of numbered mentions), and text with these numbered mentions marked.\newline{}- Your task is: To determine if all mentions in the cluster clearly point to the same entity. If all mentions point to the same entity, output ``Yes''; if at least one mention points to a different entity, output ``No''; if there is insufficient information to determine, output ``Pending''.\newline{}\newline{}\#\#\# Key Definitions\newline{}- \textbf{Mention}: Refers to a specific occurrence in the text of one or more consecutive words used to refer to an entity (person/object/place/organization/concept, etc.), including pronouns, noun phrases, or descriptive spans. A mention must be a sequence of consecutive words, and usually does not include irrelevant punctuation (such as periods, question marks, or commas), unless the punctuation is an integral part of the entity's name (e.g., ``U.S.A.'').\newline{}- \textbf{Coreference Task}: Group mentions at different positions in the text that point to the same entity into one cluster.\newline{}- \textbf{Mention Numbering Format}: Each mention is assigned a unique number in the format ``\#X'', where X is a numeral (e.g., ``\#1: you'' indicates that the mention ``you'' is numbered \#1).\newline{}- \textbf{Mention Annotation Format in Text}: In text with numbered mentions, the start of each mention is marked with ``[(\#X)'' and the end with ``](\#X)'' (where X is a numeral), with no space between the markers and adjacent words (e.g., ``[(\#1)you](\#1)'' represents the mention ``you'' numbered \#1). This ensures the location of each mention instance is unique, avoiding confusion between mentions with the same form.\newline{}\newline{}\#\#\# Checking Criteria\newline{}- \textbf{Determine ``Yes''}: When all mentions clearly refer to the same entity in context.\newline{}- \textbf{Determine ``No''}: When at least one mention in the cluster clearly refers to a different entity.\newline{}- \textbf{Determine ``Pending''}: When the context does not allow confirmation of coreference relationships for all mentions.\newline{}\newline{}\#\#\# Output Format Requirements\newline{}- First, output the reason for the judgment (a brief analysis within 50 words).\newline{}- Then, output the judgment result, which must be: ``Yes'', ``No'', or ``Pending''.\newline{}- The output should end with the judgment result, with no extra content or explanation.\newline{}- Example output structure: ``Reason text. Yes'', ``Reason text. No'', or ``Reason text. Pending'' \\
     \\
    \rowcolor{Gray}
    \#\#\# Input to be Checked\newline{}
    Original text: \newline{}
SOUTH AFRICA FREED the ANC 's Sisulu and seven other political prisoners . Thousands of supporters , many brandishing flags of the outlawed African National Congress , gave the anti-apartheid activists a tumultuous reception upon their return to black townships across the country . Most of those freed had spent at least 25 years in prison . The 77 - year - old Sisulu , sentenced to life in 1964 along with black nationalist Nelson Mandela for plotting to overthrow the government , said equality for blacks in South Africa was in reach . The releases , announced last week by President de Klerk , were viewed as Pretoria 's tacit legalization of the ANC . Mandela , considered the most prominent leader of the ANC , remains in prison . 
\newline{}\newline{}
Coreference cluster result (with a set of numbered mentions): \newline{}
\#1:SOUTH AFRICA, \#2:FREED, \#3:the ANC 's, , \#4:the outlawed African National Congress, \#5:the country, \#6:the government, \#7:South Africa, \#8:The releases , announced last week by President de Klerk ,, \#9:Pretoria 's, \#10:the ANC, \#11:the ANC
\newline{}\newline{}
Text with these numbered mentions: \newline{}
[(\#1)SOUTH AFRICA](\#1) [(\#2)FREED](\#2) [(\#3)the ANC 's](\#3) Sisulu and seven other political prisoners . Thousands of supporters , many brandishing flags of [(\#4)the outlawed African National Congress](\#4) , gave the anti-apartheid activists a tumultuous reception upon their return to black townships across [(\#5)the country](\#5) . Most of those freed had spent at least 25 years in prison . The 77 - year - old Sisulu , sentenced to life in 1964 along with black nationalist Nelson Mandela for plotting to overthrow [(\#6)the government](\#6) , said equality for blacks in [(\#7)South Africa](\#7) was in reach . [(\#8)The releases , announced last week by President de Klerk ,](\#8) were viewed as [(\#9)Pretoria 's](\#9) tacit legalization of [(\#10)the ANC](\#10) . Mandela , considered the most prominent leader of [(\#11)the ANC](\#11) , remains in prison .
   \\
    LLM Output:\newline{}
    The cluster includes mentions of South Africa (country) and the ANC (organization), which are distinct entities. Mentions like \#1 (SOUTH AFRICA), \#5 (the country), \#7 (South Africa), and \#9 (Pretoria's, referring to South Africa's government) point to the country. Mentions \#3 (the ANC's), \#4 (the African National Congress), \#10, and \#11 refer to the ANC. These are different entities.  
No
    \\
    \bottomrule
    \end{tabular}%
    }
    \caption{Example 4 of LLM coreference Checker-Splitter with the incorrect coreference cluster (\textbf{coreference checker}).}
  \label{tab:example4_cluster_checker}%
\end{table*}%

\begin{table*}[htbp]
  \centering
  \resizebox{2.0\columnwidth}{!}{
    \begin{tabular}{p{88.065em}}
    \toprule
    Your task is to regroup a coreference cluster.\newline{}\newline{}\#\#\# Task Description\newline{}- Input includes: A original text, a coreference cluster result (with a set of numbered mentions), and text with these numbered mentions marked.\newline{}- Your task is: The input coreference cluster has been identified as incorrect (not all mentions point to the same entity). You need to regroup these mentions based on the text context, ensuring each new group's mentions refer to the same entity, and different groups' mentions refer to different entities. The output must include the reasons for regrouping and the regrouping results.\newline{}\newline{}\#\#\# Key Definitions\newline{}- \textbf{Mention}: Refers to a specific occurrence in the text of one or more consecutive words used to refer to an entity (person/object/place/organization/concept, etc.), including pronouns, noun phrases, or descriptive spans. A mention must be a sequence of consecutive words, and usually does not include irrelevant punctuation (such as periods, question marks, or commas), unless the punctuation is an integral part of the entity's name (e.g., ``U.S.A.'').\newline{}- \textbf{Coreference Task}: Group mentions at different positions in the text that point to the same entity into one cluster.\newline{}- \textbf{Mention Numbering Format}: Each mention is assigned a unique number in the format``\#X'', where X is a numeral (e.g., ``\#1: you'' indicates that the mention ``you'' is numbered \#1).\newline{}- \textbf{Mention Annotation Format in Text}: In text with numbered mentions, the start of each mention is marked with ``[(\#X)'' and the end with ``](\#X)'' (where X is a numeral), with no space between the markers and adjacent words (e.g., ``[(\#1)you](\#1)'' represents the mention ``you'' numbered \#1).  This ensures the location of each mention instance is unique, avoiding confusion between mentions with the same form.\newline{}- \textbf{Mention Annotation Method}: All mentions in the input are marked in the above format. Only these numbered mentions can be used for grouping. Unmarked parts of the text only provide context.\newline{}\newline{}\#\#\# Regrouping Criteria\newline{}Strictly follow these rules for regrouping:\newline{}- \textbf{Intra-group Consistency}: All mentions in a new group must clearly refer to the same entity.\newline{}- \textbf{Inter-group Exclusivity}: Mentions in different groups must refer to different entities.\newline{}- \textbf{Completeness}: The regrouping must include all mentions from the original cluster without omission.\newline{}- \textbf{Conflict Resolution}: If a mention's affiliation is unclear, place it in a new group.\newline{}- \textbf{Scope Limitation}: Only process mentions in the input cluster; do not introduce other unmarked mentions from the text.\newline{}- \textbf{Nested Mention Handling}: For nested mentions (one mention containing another), determine their affiliation based on context: group them together if they refer to the same entity, otherwise separate them.\newline{}- \textbf{Failure Handling}: If correction is impossible (e.g., insufficient context or ambiguous mention reference), return the reason for failure. Otherwise, output the reasons and results.\newline{}\newline{}\#\#\# Output Format Requirements\newline{}- First, provide a brief analysis (within 100 words) of the original cluster's error and the regrouping basis.\newline{}- Then, output the regrouped mention numbers. The format is comma-separated groups, each group expressed as: [\#list of numbers] (e.g., [\#1,\#2,\#4], [\#3,\#5]). Numbers must be in ascending order and only from the original cluster.\newline{}- The output must end with the grouping result, with no extra content or explanation.\newline{}- Example output structure: ``Reason text. [\#1, ...], [\#X, ..], ...''.\newline{}- If correction fails, output ``Correction failed:'' followed by the reason (e.g., ``Correction failed: Insufficient context to determine the entity reference of \#3''), then end the output. \\
     \\
    \rowcolor{Gray}
    \#\#\# Input to be Regrouped\newline{}
    Original text: \newline{}
SOUTH AFRICA FREED the ANC 's Sisulu and seven other political prisoners . Thousands of supporters , many brandishing flags of the outlawed African National Congress , gave the anti-apartheid activists a tumultuous reception upon their return to black townships across the country . Most of those freed had spent at least 25 years in prison . The 77 - year - old Sisulu , sentenced to life in 1964 along with black nationalist Nelson Mandela for plotting to overthrow the government , said equality for blacks in South Africa was in reach . The releases , announced last week by President de Klerk , were viewed as Pretoria 's tacit legalization of the ANC . Mandela , considered the most prominent leader of the ANC , remains in prison . 
\newline{}\newline{}
Coreference cluster result (with a set of numbered mentions): \newline{}
\#1:SOUTH AFRICA, \#2:FREED, \#3:the ANC 's, , \#4:the outlawed African National Congress, \#5:the country, \#6:the government, \#7:South Africa, \#8:The releases , announced last week by President de Klerk ,, \#9:Pretoria 's, \#10:the ANC, \#11:the ANC
\newline{}\newline{}
Text with these numbered mentions: \newline{}
[(\#1)SOUTH AFRICA](\#1) [(\#2)FREED](\#2) [(\#3)the ANC 's](\#3) Sisulu and seven other political prisoners . Thousands of supporters , many brandishing flags of [(\#4)the outlawed African National Congress](\#4) , gave the anti-apartheid activists a tumultuous reception upon their return to black townships across [(\#5)the country](\#5) . Most of those freed had spent at least 25 years in prison . The 77 - year - old Sisulu , sentenced to life in 1964 along with black nationalist Nelson Mandela for plotting to overthrow [(\#6)the government](\#6) , said equality for blacks in [(\#7)South Africa](\#7) was in reach . [(\#8)The releases , announced last week by President de Klerk ,](\#8) were viewed as [(\#9)Pretoria 's](\#9) tacit legalization of [(\#10)the ANC](\#10) . Mandela , considered the most prominent leader of [(\#11)the ANC](\#11) , remains in prison .
   \\
    LLM Output:\newline{}
    The original cluster incorrectly groups distinct entities: country (South Africa), organization (ANC), government (Pretoria), and non-entity actions/events (FREED, releases). Regrouping based on entity consistency:  
- Mentions \#1 (SOUTH AFRICA), \#5 (the country), \#7 (South Africa) refer to the nation.  
- \#3 (the ANC's), \#4 (the outlawed African National Congress), \#10 (the ANC), \#11 (the ANC) refer to the ANC organization.  
- \#6 (the government) and \#9 (Pretoria's) refer to the governing authority.  
- \#2 (FREED) is an action, not an entity.  
- \#8 (The releases...) describes an event, not a coreferring entity.  
[\#1,\#5,\#6,\#7,\#9], [\#2,\#8], [\#3,\#4,\#10,\#11]  
    \\
    \bottomrule
    \end{tabular}%
    }
    \caption{Example 4 of LLM coreference Checker-Splitter with the incorrect coreference cluster (\textbf{coreference corrector}).}
  \label{tab:example4_cluster_corrector}%
\end{table*}%

\section{Full Pronouns List}
\label{app:full_pronouns_list}
We list all the pronouns in Table~\ref{tab:full_pronouns_list}. And we illustrate the text segmentation strategies and LBM in Fig.~\ref{pic2_lbm:app}

\begin{table}[htbp]
  \centering
  \resizebox{1.0\columnwidth}{!}{
    \begin{tabular}{p{20.455em}}
    \toprule
    i , me , my , mine , myself , you , your , yours , yourself , yourselves , he , him , his , himself , she , her , hers , herself , it , its , itself , we , us , our , ours , ourselves , they , them , their , themselves , that , this  \\
    \bottomrule
    \end{tabular}%
    }
    \caption{Full pronouns list. Note that in the specific experimental process, ``it'' should be removed from full pronouns list because ``it'' may lack referential significance. }
  \label{tab:full_pronouns_list}%
\end{table}%

\section{Complete Experimental Settings}
\label{app:com_exp_set}
\textbf{Datasets.} We train and evaluate the performance of our proposed methods on two widely used CR datasets: OntoNotes~\cite{pradhan-etal-2012-conll} and LitBank~\cite{bamman-etal-2020-annotated}. 
Specifically, OntoNotes, originating from the CoNLL-2012 shared task, serves as the de facto standard for evaluating CR methods. It encompasses seven text genres, including full-length documents such as newswire articles, broadcast news, magazines, web text, and testament passages, as well as multi-speaker transcripts like broadcast conversations and telephone conversations. LitBank is frequently employed for evaluating CR on long documents and comprises $100$ literary works. 
Note that OntoNotes and LitBank follow different annotation guidelines: OntoNotes does not annotate singleton clusters (i.e., single-mention clusters), whereas LitBank provides such annotations. 
To assess the generalization capability of methods, models trained on OntoNotes are also tested on the out-of-domain dataset WikiCoref~\cite{ghaddar-langlais-2016-wikicoref}. 
WikiCoref only contains $30$ Wikipedia articles as its test set, with some texts reaching lengths of up to $9869$ words. 
We detail these datasets in Table~\ref{tab:Dataset_statistics_app}.

\begin{table}[htbp]
  \centering
  \resizebox{1.0\columnwidth}{!}{
    \begin{tabular}{l|ccccc}
    \hline
    Datasets & \#Train & \#Val & \#Test & Avg. W & Avg. M  \\
    \hline
    OntoNotes & 2802  & 343   & 348   & 467   & 56      \\
    LitBank & 80    & 10    & 10    & 2105  & 291     \\
    WikiCoref & -     & -     & 30    & 1996  & 230    \\
    \hline
    \end{tabular}%
    }
    \caption{Dataset statistics: number of documents in each dataset split (\#Train/Val/Test), average number of words (i.e., Avg. W) and mentions (i.e., Avg. M) per document.}
  \label{tab:Dataset_statistics_app}%
\end{table}%

\textbf{Baselines.} To gauge the effectiveness of our methods, we follow~~\cite{martinelli-etal-2024-maverick} by adopting DeBERTa$_{\text{large}}$~\cite{he2021deberta} as the base model for \ourmethodfirst. 
Meanwhile, we choose three latest and most robust LLMs for \ourmethod: Qwen2.5-72B-Instruct~\cite{qwen2.5}, DeepSeek-R1-250528~\cite{deepseekai2025deepseekr1incentivizingreasoningcapability} and GPT-4~\cite{le2023large}. For clarity, we denote \ourmethod~with different LLMs as \textbf{\ourmethod$_{\text{qwen/dpsk/gpt4}}$}. 
Also, we compare \ourmethod~and \ourmethodfirst~against numerous baselines: 1) \textbf{Supervised neural methods} including c2f-coref~\cite{joshi2020spanbert}, ICoref~\cite{xia-etal-2020-incremental}, CorefQA~\cite{wu-etal-2020-corefqa}, s2e-coref~\cite{kirstain2021coreference}, longdoc~\cite{toshniwal2021generalization}, wl-coref~\cite{dobrovolskii-2021-word}, f-coref~\cite{otmazgin2022f}, LingMess~\cite{otmazgin2022lingmess}, Dual-cache~\cite{guo-etal-2023-dual}, Maverick$_{\text{mes}}$~\cite{martinelli-etal-2024-maverick}; 2) \textbf{Generative methods} containing ASP~\cite{liu-etal-2022-autoregressive}, Link-Append~\cite{bohnet-etal-2023-coreference}, seq2seq~\cite{zhang-etal-2023-seq2seq}; 3) \textbf{Large Language Models} comprising InstructGPT~\cite{le2023large}, GPT-3.5-turbo~\cite{le2023large, zhu-etal-2025-llmlink}, GPT-4~\cite{le2023large} and LLMLink~\cite{zhu-etal-2025-llmlink}. 
Notably, for fairness, our methods align with the text encoding strategy of Maverick$_{\text{mes}}$ in comparative experiments, without considering the LBM module. The utility of LBM will be thoroughly investigated in the ablation study.

\textbf{Metrics.}
In our experiments, coreference performance across different methods is evaluated using the MUC~\cite{vilain1995model}, B$^3$~\cite{bagga1998algorithms}, and CEAF$_{\phi_4}$~\cite{luo2005coreference} metrics, with Precision (P), Recall (R), and F1 scores reported for each. The overall performance of a method is measured by the CoNLL-F1 score, computed as the average of the F1 scores from these three metrics (abbreviated as Avg.F1). To further analyze model capabilities, certain ablation studies also report Precision (P), Recall (R), and F1 scores for Mention Detection. 

\textbf{Configurations.}
In all experiments, unless otherwise specified, we default to setting $L_{\text{max}}=30$ for HyMR, $\eta_1=\eta_2=0.6$, and $\rho=1e-3$. Meanwhile, we employ the Adafactor optimizer~\cite{shazeer2018adafactor} for model training, setting the learning rate for the DeBERTa$_{\text{large}}$ to 2e-5 and for the remaining model layers to 3e-4. 
All experiments are implemented using the PyTorch-Lightning framework. Each run is executed on a single RTX 4090 GPU with 24GB of VRAM. We have eight such GPUs available, allowing multiple experiments to be conducted concurrently. For fairness, neither data nor model parallelism is employed in any experiment.
During training, we accumulate gradients every four steps and set the gradient clipping threshold at $1.0$. A linear learning rate scheduler is adopted, incorporating a warm-up phase covering $10\%$ of all training steps. To monitor performance, a validation evaluation is performed every one epoch. The final model is selected based on Avg.F1 in the validation set, with an early stopping patience of $30$. For the LLMs utilized, we access them remotely via their API interfaces, uniformly setting the temperature parameter to $0$.

\section{Additional Experimental Results}
\label{app:add_exp_res}
\begin{figure*}[ht]
  \centering
  \includegraphics[width=1.0\linewidth]{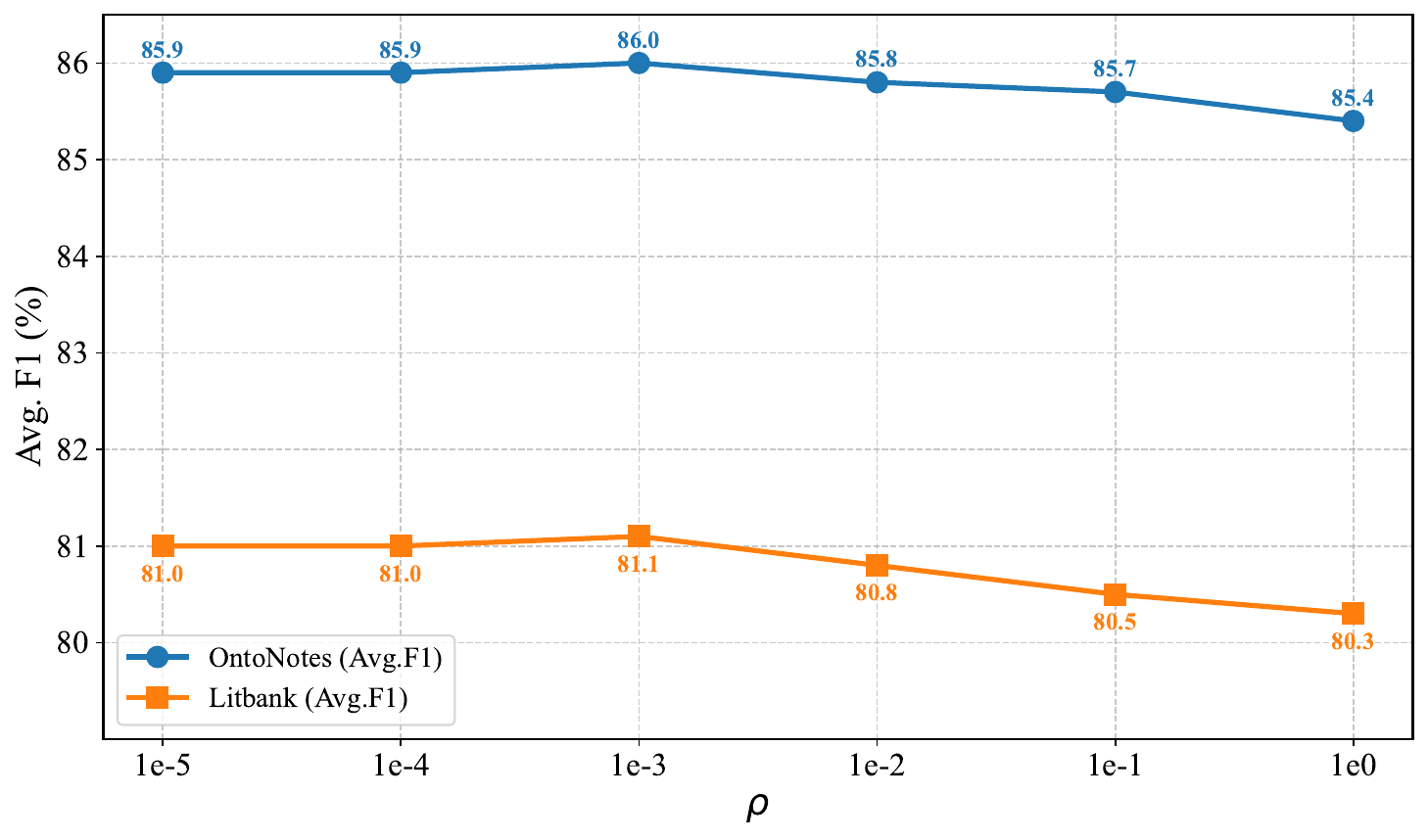}
  \caption{Avg.F1 (\%) from ImCoref-CeS$_\text{gpt4}$ with varying $\rho$ .} 
  \label{pic_rho:app}
\end{figure*}

In this section, we report additional experimental results, as detailed in Tables~\ref{tab:Per_com_ontonotes_app}-\ref{tab:ceco_app}. Meanwhile, we utilize ImCoref-CeS$_\text{gpt4}$ to report the performance across different $\rho$ values on OntoNotes and LitBank, as shown in Fig.~\ref{pic_rho:app}.

\begin{table*}[t]
  \centering 
  \resizebox{2.0\columnwidth}{!}{
    \begin{tabular}{ll|ccc|ccc|ccc|c}
    \hline
    \multicolumn{1}{c}{\multirow{2}[2]{*}{Methods}} & \multicolumn{1}{c|}{\multirow{2}[2]{*}{Base Encoders}} & \multicolumn{3}{c|}{MUC} & \multicolumn{3}{c|}{B$^3$} & \multicolumn{3}{c|}{CEAF$_{\phi_4}$} & \multirow{2}[2]{*}{Avg.F1} \\
          &       & P     & R     & F1    & P     & R     & F1    & P     & R     & F1    &  \\
    \hline
    \rowcolor{Gray}\multicolumn{12}{l}{\textbf{Supervised neural methods}} \\
    c2f-coref & SpanBERT$_{\text{large}}$ & 85.8  & 84.8  & 85.3  & 78.3  & 77.9  & 78.1  & 76.4  & 74.2  & 75.3  & 79.6 \\
    Icoref & SpanBERT$_{\text{large}}$ & 85.7  & 84.8  & 85.3  & 78.1  & 77.5  & 77.8  & 76.3  & 74.1  & 75.2  & 79.4 \\
    CorefQA & SpanBERT$_{\text{large}}$ & \textbf{88.6}  & 87.4  & 88.0  & 82.4  & 82.0  & 82.2  & 79.9  & 78.3  & 79.1  & 83.1$^\diamond$ \\
    s2e-coref & LongFormer$_{\text{large}}$ & 86.5  & 85.1  & 85.8  & 80.3  & 77.9  & 79.1  & 76.8  & 75.4  & 76.1  & 80.3 \\
    longdoc & LongFormer$_{\text{large}}$ & 85.5  & 85.1  & 85.3  & 78.7  & 77.3  & 78.0  & 74.2  & 76.5  & 75.3  & 79.6 \\
    wl-coref & RoBERTa$_{\text{large}}$ & 84.9  & 87.9  & 86.3  & 77.4  & 82.6  & 79.9  & 76.1  & 77.1  & 76.6  & 81.0 \\
    f-coref & DistilRoBERTa & 85.0  & 83.9  & 84.4  & 77.6  & 75.5  & 76.6  & 74.7  & 74.3  & 74.5  & 78.5$^\diamond$ \\
    LingMess & LongFormer$_{\text{large}}$ & 88.1  & 85.1  & 86.6  & 82.7  & 78.3  & 80.5  & 78.5  & 76.0  & 77.3  & 81.4 \\
    Dual-cache & LongFormer$_{\text{large}}$ & -     & -     & 86.3  & -     & -     & 80.3  & -     & -     & 76.8  & 81.1 \\
    Maverick$_{\text{mes}}$ & DeBERTa$_{\text{large}}$ & 87.5  & \textbf{88.5}  & 88.0  & 82.2  & 83.5  & 82.8  & 80.4  & 79.3  & 79.9  & 83.6 \\
    ImCoref(\textbf{ours}) & DeBERTa$_{\text{large}}$ & \textbf{88.2}  & 88.3  & \textbf{88.2}  & \textbf{83.4}  & \textbf{83.8}  & \textbf{83.6}  & \textbf{81.9}  & \textbf{80.1}  & \textbf{81.0}  & \textbf{84.3} \\
    \hline
    \rowcolor{Gray}\multicolumn{12}{l}{\textbf{Generative methods}} \\
    ASP   & FLAN-T5$_{\text{XXL}}$ & 86.1  & 88.4  & 87.2  & 80.2  & 83.2  & 81.7  & 78.9  & 78.3  & 78.6  & 82.5 \\
    Link-Append & mT5$_{\text{XXL}}$ & \textbf{87.4}  & 88.3  & \textbf{87.8}  & \textbf{81.8}  & 83.4  & \textbf{82.6}  & \textbf{79.1}  & 79.9  & \textbf{79.5}  & \textbf{83.3} \\
    seq2seq & T0$_{\text{XXL}}$ & 86.1  & \textbf{89.2}  & 87.6  & 80.6  & \textbf{84.3}  & 82.4  & 78.9  & \textbf{80.1}  & 79.5  & 83.2 \\
    \hline
    \rowcolor{Gray}\multicolumn{12}{l}{\textbf{Large Language Models}} \\
    \multirow{2}[0]{*}{InstructGPT} & \multirow{2}[0]{*}{API} & 71.1  & 69.7  & 70.4  & 58.1  & 58.6  & 58.4  & 60.6  & 45.1  & 51.7  & 60.1 \\
          &       & 89.6  & 88.9  & 89.2  & 76.0  & 89.2  & 79.4  & \textbf{84.8}  & 65.2  & 73.7  & 80.8$^\star$ \\
    \hdashline
    \multirow{2}[0]{*}{GPT-3.5-turbo} & \multirow{2}[0]{*}{API} & 67.3  & 66.5  & 66.9  & 54.3  & 56.8  & 55.5  & 43.9  & 49.5  & 46.5  & 56.3 \\
          &       & 88.2  & 84.4  & 86.2  & 79.3  & 79.3  & 79.3  & 65.6  & 71.2  & 68.3  & 77.9$^\star$ \\
    \hdashline
    \multirow{2}[1]{*}{GPT-4} & \multirow{2}[1]{*}{API} & 73.9  & 73.5  & 73.7  & 60.8  & 64.7  & 62.7  & 49.3  & 55.7  & 52.3  & 62.9 \\
          &       & \textbf{93.8}  & \textbf{93.7}  & \textbf{93.7}  & \textbf{86.5}  & \textbf{91.1}  & \textbf{88.8}  & 83.5  & \textbf{82}    & \textbf{82.8}  & \textbf{88.4}$^\star$ \\
    \hline
    \rowcolor{Gray}\multicolumn{12}{l}{\textbf{Supervised neural methods+Large Language Models (ours)}} \\
    ImCoref-CeS$_{\text{qwen}}$ & DeBERTa$_{\text{large}}$+API & 90.2  & 89.1  & 89.6  & 84.8  & 83.5  & 84.1  & 82.9  & 79.2  & 81.0  & 84.9 \\
    ImCoref-CeS$_{\text{dpsk}}$ & DeBERTa$_{\text{large}}$+API & 92.1  & 89.5  & 90.8  & 85.4  & 83.9  & 84.6  & 83.6  & 79.7  & 81.6  & 85.7 \\
    ImCoref-CeS$_{\text{gpt4}}$ & DeBERTa$_{\text{large}}$+API & \textbf{92.4}  & \textbf{90.1}  & \textbf{91.2}  & \textbf{85.7}  & \textbf{84.1}  & \textbf{84.9}  & \textbf{83.8}  & \textbf{79.9}  & \textbf{81.8}  & \textbf{86.0} \\
    \hline
    \end{tabular}%
    }
  \caption{Performance comparison (\%) of different methods on OntoNotes. Note that ($\diamond$) indicates models trained with additional resources, while ($\star$) signifies that their coreference clustering was performed using provided gold mentions.}
  \label{tab:Per_com_ontonotes_app}%
\end{table*}%

\begin{table*}[htbp]
  \centering
    \begin{tabular}{lccc|ccc|ccc|c}
    \hline
    \multicolumn{1}{c}{\multirow{2}[2]{*}{Methods}} & \multicolumn{3}{c|}{MUC} & \multicolumn{3}{c|}{B$^3$} & \multicolumn{3}{c|}{CEAF$_{\phi_4}$} & \multirow{2}[2]{*}{Avg.F1} \\
          & P     & R     & F1    & P     & R     & F1    & P     & R     & F1    &  \\
    \hline
    longdoc & 90.8  & 85.7  & \textbf{88.2}  & 80.0    & 72.1  & 75.9  & 65.1  & 66.0    & 65.5  & 76.5 \\
    Dual-cache & -     & -     & \textbf{88.2}  & -     & -     & 79.2  & -     & -     & 71.0    & 79.5 \\
    Maverick$_{\text{mes}}$ & 94.1  & 79.9  & 86.5  & 82.2  & 75.7  & 78.8  & 59.1  & 85.2  & 69.8  & 78.4 \\
    ImCoref(ours) & 93.4  & 83.0    & 87.9  & 79.1  & 79.9  & \textbf{79.5}  & 62.8  & 83.3  & \textbf{71.6}  & \textbf{79.7} \\
    \hline
    seq2seq & -     & -     & -     & -     & -     & -     & -     & -     & -     & 77.3 \\
    \hline
    GPT-3.5-turbo & -     & -     & -     & -     & -     & -     & -     & -     & -     & 75.3 \\
    LLMLink & -     & -     & -     & -     & -     & -     & -     & -     & -     & 81.5$^\star$ \\
    \hline
    ImCoref-CeS$_{\text{qwen}}$ & 95.2  & 83.1  & 88.7  & 81.3  & 79.2  & 80.2  & 66.5  & 79.1  & 72.3  & 80.4 \\
    ImCoref-CeS$_{\text{dpsk}}$ & 95.3  & 83.6  & \textbf{89.1}  & 82.8  & 81.1  & \textbf{81.9}  & 69.4  & 80.5  & \textbf{74.5}  & \textbf{81.8} \\
    ImCoref-CeS$_{\text{gpt4}}$ & 94.8  & 83.5  & 88.8  & 82.4  & 80.6  & 81.5  & 67.2  & 79.7  & 72.9  & 81.1 \\
    \hline
    \end{tabular}%
    \caption{Performance comparison (\%) of different methods on LitBank. Note that ($\star$)  indicates that it is required to train multiple LLMs to perform coreference task.}
  \label{tab:Per_com_litbank_app}%
\end{table*}%

\begin{table*}[htbp]
  \centering
    \begin{tabular}{lccc|ccc|ccc|c}
    \hline
    \multicolumn{1}{c}{\multirow{2}[2]{*}{Methods}} & \multicolumn{3}{c|}{MUC} & \multicolumn{3}{c|}{B$^3$} & \multicolumn{3}{c|}{CEAF$_{\phi_4}$} & \multirow{2}[2]{*}{Avg.F1} \\
          & P     & R     & F1    & P     & R     & F1    & P     & R     & F1    &  \\
    \hline
    longdoc & -     & -     & -     & -     & -     & -     & -     & -     & -     & 60.1  \\
    Dual-cache & -     & -     & 72.1  & -     & -     & 62.1  & -     & -     & 54.7  & 63.0  \\
    LingMess & -     & -     & -     & -     & -     & -     & -     & -     & -     & 62.6  \\
    Maverick$_{\text{mes}}$ & 87.7  & 76.0  & \textbf{81.5}  & 65.1  & 65.7  & 65.4  & 45.8  & 64.3  & 53.5  & 66.8  \\
    ImCoref(ours) & 87.4  & 74.9  & 80.6  & 68.6  & 64.8  & \textbf{66.7}  & 50.0  & 62.5  & \textbf{55.6}  & \textbf{67.6}  \\
    \hline
    InstructGPT & -     & -     & -     & -     & -     & -     & -     & -     & -     & 72.9  \\
    GPT-3.5-turbo & -     & -     & -     & -     & -     & -     & -     & -     & -     & 70.8  \\
    \hline
    ImCoref-CeS$_{\text{qwen}}$ & 83.8  & 79.8  & 81.8  & 66.9  & 67.7  & 67.3  & 71.0  & 59.9  & 65.0  & 71.3  \\
    ImCoref-CeS$_{\text{dpsk}}$ & 85.9  & 79.1  & 82.4  & 70.3  & 68.1  & 69.2  & 72.8  & 59.6  & 65.5  & 72.4  \\
    ImCoref-CeS$_{\text{gpt4}}$ & 85.8  & 81.5  & \textbf{83.6}  & 68.9  & 70.8  & \textbf{69.8}  & 73.4  & 60.1  & \textbf{66.1}  & \textbf{73.2}  \\
    \hline
    \end{tabular}%
    \caption{Performance comparison (\%) of different methods on WikiCoref.}
  \label{tab:Per_com_wikiCoref_app}%
\end{table*}%

\begin{table*}[htbp]
  \centering
  \resizebox{2.0\columnwidth}{!}{
    \begin{tabular}{lc|ccc|ccc|ccc|c|c|cc}
    \hline
    \multicolumn{1}{c}{\multirow{2}[2]{*}{Methods}} & \multirow{2}[2]{*}{Parameters} & \multicolumn{3}{c|}{MUC} & \multicolumn{3}{c|}{B$^3$} & \multicolumn{3}{c|}{CEAF$_{\phi_4}$} & \multirow{2}[2]{*}{Avg. F1} & Training & \multicolumn{2}{c}{Inference} \\
          &       & P     & R     & F1    & P     & R     & F1    & P     & R     & F1    &       & Time  & Time  & Mem. \\
    \hline
    \rowcolor{Gray}\multicolumn{12}{l}{OntoNotes}                                                                & \multicolumn{1}{c}{} &       &  \\
    baseline & 531M  & 88.2  & 88.3  & 88.2  & 83.4  & 83.8  & 83.6  & 81.9  & 80.1  & 81.0  & 84.3  & 13.0h & 15.0s & 4.05  \\
    \hdashline
    -w. Independent & 531M  & 86.1  & 86.3  & 86.2  & 80.0  & 79.3  & 79.6  & 80.4  & 76.4  & 78.3  & 81.4  & 13.1h & 15.2s & 4.05  \\
    -w. Overlap & 531M  & 87.1  & 86.3  & 86.7  & 82.3  & 79.0  & 80.6  & 79.4  & 78.4  & 78.9  & 82.1  & 13.8h & 17.2s & 4.12  \\
    -w. LBM-FC & 534M  & 87.9  & 86.9  & 87.4  & 83.2  & 81.1  & 82.1  & 80.3  & 79.0  & 79.6  & 83.1  & 13.5h & 16.0s & 4.07  \\
    -w. LBM-MHA & 540M  & 86.2  & 89.3  & 87.7  & 80.8  & 83.8  & 82.3  & 81.9  & 80.0  & 80.9  & 83.6  & 13.6h & 16.3s & 4.15  \\
    \hline
    \rowcolor{Gray}\multicolumn{12}{l}{Litbank}                                                                  & \multicolumn{1}{c}{} &       &  \\
    baseline & 531M  & 93.4  & 83.0  & 87.9  & 79.1  & 79.9  & 79.5  & 62.8  & 83.3  & 71.6  & 79.2  & 3.00h & 4.0s  & 3.87  \\
    \hdashline
    -w. Independent & 531M  & 92.7  & 80.4  & 86.1  & 77.2  & 76.7  & 76.9  & 60.3  & 74.2  & 66.5  & 76.5  & 3.20h & 4.1s  & 3.87  \\
    -w. Overlap & 531M  & 92.4  & 83.1  & 87.5  & 77.5  & 79.2  & 78.3  & 60.0  & 82.2  & 69.4  & 78.4  & 3.80h & 4.8s  & 3.94  \\
    -w. LBM-FC & 534M  & 93.3  & 82.9  & 87.8  & 77.2  & 79.1  & 78.1  & 62.7  & 82.3  & 71.2  & 79.0  & 3.25h & 4.2s  & 3.89  \\
    -w. LBM-MHA & 540M  & 93.1  & 83.0  & 87.8  & 78.4  & 79.3  & 78.8  & 61.4  & 83.0  & 70.6  & 79.1  & 3.30h & 4.4s  & 3.91  \\
    \hline
    \end{tabular}%
    }
    \caption{Performance comparison (\%) of different concatenation strategies on OntoNotes and LitBank datasets.}
  \label{tab:concatenation_strategies_app}
\end{table*}%

\begin{table*}[htbp]
  \centering
  \resizebox{2.0\columnwidth}{!}{
    \begin{tabular}{lccc|ccc|ccc|c|c|cc}
    \hline
    \multirow{2}[2]{*}{} & \multicolumn{3}{c|}{MUC} & \multicolumn{3}{c|}{B$^3$} & \multicolumn{3}{c|}{CEAF$_{\phi_4}$} & \multirow{2}[2]{*}{Avg.F1} & Training & \multicolumn{2}{c}{Inference} \\
          & P     & R     & F1    & P     & R     & F1    & P     & R     & F1    &       & Time  & Time  & Mem. \\
    \hline
     \rowcolor{Gray}\multicolumn{14}{l}{OntoNotes} \\
    10    & 88.4  & 87.2  & 87.8  & 83.4  & 82.8  & 83.1  & 82.3  & 77.8  & 80.0  & 83.6  & 12.4h & 14.1s & 3.84 \\
    20    & 88.0  & 88.6  & 88.3  & 81.9  & 85.2  & 83.5  & 82.4  & 79.0  & 80.7  & 84.2  & 12.8h & 14.5s & 3.97 \\
    30    & 88.2  & 88.3  & 88.2  & 83.4  & 83.8  & 83.6  & 81.9  & 80.1  & 81.0  & 84.3  & 13.0h & 15.0s & 4.05 \\
    40    & 88.4  & 88.2  & 88.3  & 83.3  & 83.7  & 83.5  & 83.1  & 79.0  & 81.0  & 84.3  & 13.7h & 15.2s & 4.16 \\
    EOS   & 88.7  & 87.9  & 88.3  & 84.7  & 83.8  & 84.2  & 82.2  & 78.8  & 80.5  & 84.3  & 14.3h & 15.9s & 4.28 \\
    \hline
     \rowcolor{Gray}\multicolumn{14}{l}{Litbank} \\
    10    & 91.3  & 84.6  & 87.8  & 76.4  & 80.0  & 78.2  & 60.6  & 82.6  & 69.9  & 78.6  & 2.50h  & 3.6s  & 3.81 \\
    20    & 91.4  & 83.3  & 87.2  & 80.8  & 79.1  & 79.9  & 62.7  & 80.7  & 70.6  & 79.2  & 2.70h  & 3.8s  & 3.85 \\
    30    & 93.4  & 83.0  & 87.9  & 79.1  & 79.9  & 79.5  & 62.8  & 83.3  & 71.6  & 79.7  & 3.00h  & 4.0s  & 3.87 \\
    40    & 91.6  & 84.8  & 88.1  & 78.8  & 79.7  & 79.2  & 64.1  & 80.7  & 71.4  & 79.6  & 3.10h  & 4.3s  & 3.88 \\
    EOS   & 92.3  & 84.2  & 88.1  & 79.6  & 79.7  & 79.6  & 62.5  & 81.9  & 70.9  & 79.5  & 3.40h  & 4.7s  & 3.91 \\
    \hline
    \end{tabular}%
    }
    \caption{Performance comparison (\%) of varying $L_{\text{max}}$ in HyMR on OntoNotes and LitBank datasets.}
  \label{tab:lmax_app}%
\end{table*}%

\begin{table*}[htbp]
  \centering
  \resizebox{2.0\columnwidth}{!}{
    \begin{tabular}{lccc|ccc|ccc|ccc|c}
    \hline
    \multicolumn{1}{c}{\multirow{2}[2]{*}{Methods}} & \multicolumn{3}{c|}{Mention Metric} & \multicolumn{3}{c|}{MUC} & \multicolumn{3}{c|}{B$^3$} & \multicolumn{3}{c|}{CEAF$_{\phi_4}$} & \multirow{2}[2]{*}{Avg.F1} \\
          & P     & R     & F1    & P     & R     & F1    & P     & R     & F1    & P     & R     & F1    &  \\
    \hline
    \rowcolor{Gray}\multicolumn{14}{l}{OntoNotes} \\
    ImCoref-CeS$_{\text{gpt4}}$ & 96.3  & 94.6  & 95.4  & 92.4  & 90.1  & 91.2  & 85.7  & 84.1  & 84.9  & 83.8  & 79.9  & 81.8  & \textbf{86.0 } \\
    \hdashline
    -w/o. M-Ce & 94.6  & 94.8  & 94.9  & 90.2  & 89.1  & 89.6  & 84.3  & 83.6  & 83.9  & 82.4  & 80.6  & 81.5  & 85.0  \\
    -w/o. C-CeS & 96.3  & 94.6  & 95.4  & 88.9  & 88.5  & 88.7  & 83.7  & 83.9  & 83.8  & 81.9  & 80.3  & 81.1  & 84.5  \\
    ImCoref & 94.6  & 94.8  & 94.9  & 88.2  & 88.3  & 88.2  & 83.4  & 83.8  & 83.6  & 81.9  & 80.1  & 81.0  & 84.3  \\
    \hline
    \rowcolor{Gray}\multicolumn{14}{l}{LitBank} \\
    ImCoref-CeS$_{\text{gpt4}}$ & 94.3  & 92.1  & 93.2  & 94.8  & 83.5  & 88.8  & 82.4  & 80.6  & 81.5  & 67.2  & 79.7  & 72.9  & \textbf{81.1}  \\
    \hdashline
    -w/o. M-Ce & 92.7  & 92.3  & 92.5  & 93.8  & 83.1  & 88.1  & 80.8  & 80.3  & 80.5  & 64.8  & 82.8  & 72.7  & 80.5  \\
    -w/o. C-CeS & 94.3  & 92.1  & 93.2  & 93.5  & 83.0  & 87.9  & 79.6  & 80.2  & 79.9  & 63.5  & 82.8  & 71.9  & 79.9  \\
    ImCoref & 92.7  & 92.3  & 92.5  & 93.4  & 83.0  & 87.9  & 79.1  & 79.9  & 79.5  & 62.8  & 83.3  & 71.6  & 79.7  \\
    \hline
    \end{tabular}%
    }
    \caption{Impact of each component for LLM Checker-Splitter over OntoNotes and LitBank datasets.}
  \label{tab:ceco_app}%
\end{table*}%

\section{Error Analysis}
\label{app:error_ana}
To intuitively demonstrate the prediction quality of our proposed methods, we present coreference resolution results for a complete input text (including \textbf{Gold output}, \textbf{ImCoref Output}, and \textbf{ImCoref-CeS$_{\text{gpt4}}$ Output}). The results reveal that ImCoref exhibits erroneous mention predictions (marked in blue) alongside missed correct mentions, while also producing incorrect coreference clusters (indicated in red). In contrast, ImCoref-CeS$_{\text{gpt4}}$ effectively eliminates erroneous mentions and corrects faulty coreference clusters, ensuring the correctness of each cluster. However, due to its execution of regrouping operations, it may introduce additional erroneous coreference chains. Therefore, how to more efficiently leverage the powerful reasoning capabilities of LLMs to enhance coreference performance remains a critical research challenge requiring breakthrough solutions.

\textbf{Gold Output:}

[(\#3)The fifth angel](\#3) blew [(\#3)his](\#3) trumpet . Then [(\#21)I](\#21) saw [(\#27)a star](\#27) fall from the sky to [(\#8)the earth](\#8) . [(\#27)The star](\#27) was given the key to [(\#10)the deep hole that leads down to [(\#15)the bottomless pit](\#10)](\#15) . Then [(\#27)the star](\#27) opened [(\#10)the hole leading to [(\#15)the pit](\#10)](\#15) . [(\#7)Smoke](\#7) came up from [(\#10)the hole](\#10) like smoke from a big furnace . The sun and sky became dark because of [(\#7)the smoke from [(\#10)the hole](\#7)](\#10) . Then [(\#6)locusts](\#6) came out of [(\#7)the smoke](\#7) and went down to [(\#8)the earth](\#8) . [(\#6)They](\#6) were given the power to sting like scorpions . [(\#6)They](\#6) were told not to damage the fields of grass or any plant or tree . [(\#6)They](\#6) were to hurt [(\#28)only those who did not have [(\#20)God 's](\#20) mark on [(\#28)their](\#28) foreheads](\#28) . [(\#6)They](\#6) were not given the power to kill [(\#28)them](\#28) but only to cause [(\#28)them](\#28) [(\#33)pain](\#33) for five months -- [(\#33)pain like a person feels when stung by a scorpion](\#33) . During those days [(\#11)people](\#11) will look for [(\#25)a way to die](\#25) , but [(\#11)they](\#11) will not find [(\#25)it](\#25) . [(\#11)They](\#11) will want to die , but death will hide from [(\#11)them](\#11) . [(\#6)The locusts](\#6) looked like horses prepared for battle . On [(\#6)their](\#6) heads [(\#6)they](\#6) wore something that looked like a gold crown . [(\#6)Their](\#6) faces looked like human faces . [(\#6)Their](\#6) hair was like women 's hair . [(\#6)Their](\#6) teeth were like lions ' teeth . [(\#6)Their](\#6) chests looked like iron breastplates . The sound [(\#6)their](\#6) wings made was like the noise of many horses and chariots hurrying into battle . [(\#6)The locusts](\#6) had [(\#5)tails](\#5) with stingers like scorpions . The power [(\#6)they](\#6) had to give people pain for five months was in [(\#6)[(\#5)their](\#6) tails](\#5) . [(\#6)They](\#6) had [(\#1)a ruler , who was the angel of [(\#15)the bottomless pit](\#1)](\#15) . [(\#23)[(\#1)His](\#1) name](\#23) in Hebrew is Abaddon . In Greek [(\#23)it](\#23) is Apollyon . [(\#12)The sixth angel](\#12) blew [(\#19)[(\#12)his](\#12) trumpet](\#19) . Then [(\#21)I](\#21) heard [(\#22)a voice](\#22) coming from the horns on the four corners of the golden altar that is before [(\#20)God](\#20) . [(\#22)It](\#22) said to [(\#12)the sixth angel who had [(\#19)the trumpet](\#12)](\#19) , `` Free [(\#26)the four angels who are tied at the great river Euphrates](\#26) . '' [(\#26)These four angels](\#26) had been kept ready for this hour and day and month and year . [(\#26)The angels](\#26) were set free to kill [(\#24)a third of [(\#30)all the people on [(\#8)the earth](\#8)](\#24)](\#30) . [(\#21)I](\#21) heard how many troops on [(\#29)horses](\#29) were in [(\#26)their](\#26) army . In [(\#21)my](\#21) vision , [(\#21)I](\#21) saw [(\#29)the horses](\#29) and [(\#31)the riders on [(\#29)the horses](\#29)](\#31) . [(\#31)They](\#31) looked like this : [(\#31)They](\#31) had breastplates that were fiery red , dark blue , and yellow like sulfur . The heads of [(\#29)the horses](\#29) looked like heads of lions . [(\#29)The horses](\#29) had [(\#16)[(\#14)fire](\#14) , [(\#4)smoke](\#4) , and [(\#9)sulfur](\#9)](\#16) coming out of [(\#29)[(\#17)their](\#29) mouths](\#17) . [(\#24)A third of [(\#30)all the people on [(\#8)earth](\#8)](\#24)](\#30) were killed by [(\#16)these three plagues coming out of [(\#29)[(\#17)the horses '](\#29) mouths](\#17) : [(\#14)the fire](\#14) , [(\#4)the smoke](\#4) , and [(\#9)the sulfur](\#9)](\#16) . [(\#29)The horses '](\#29) power was in [(\#29)[(\#17)their](\#29) mouths](\#17) and also in [(\#32)[(\#29)their](\#29) tails](\#32) . [(\#32)[(\#29)Their](\#29) tails](\#32) were like snakes that have heads to bite and hurt people . [(\#18)The other people on [(\#8)earth](\#8)](\#18) were not killed by [(\#16)these plagues](\#16) . But [(\#18)these people](\#18) still did not change [(\#18)[(\#13)their](\#18) hearts](\#13) and turn away from worshiping the things [(\#18)they](\#18) had made with [(\#18)their](\#18) own hands . [(\#18)They](\#18) did not stop worshiping [(\#2)demons and idols made of gold , silver , bronze , stone , and wood](\#2) -- [(\#2)things that can not see or hear or walk](\#2) . [(\#18)They](\#18) did not change [(\#18)[(\#13)their](\#18) hearts](\#13) and turn away from killing other people or from [(\#18)their](\#18) evil magic , [(\#18)their](\#18) sexual sins , and [(\#18)their](\#18) stealing .

\textbf{ImCoref Output:}

[(\#3)The fifth angel](\#3) blew [(\#3)his](\#3) trumpet . Then [(\#21)I](\#21) saw [(\#27)a star](\#27) fall from [({\color{red} \#32}){\color{blue} the sky}]({\color{red} \#32}) to [(\#8)the earth](\#8) . [(\#27)The star](\#27) was given [({\color{red} \#32}){\color{blue} the sky}]({\color{red} \#32}) to [(\#10)the deep hole that leads down to [(\#15)the bottomless pit](\#10)](\#15) . Then [(\#27)the star](\#27) opened [(\#10)the hole leading to [(\#15)the pit](\#10)](\#15) . [(\#7){\color{blue} Smoke}](\#7) came up from [(\#10)the hole](\#10) like smoke from a big furnace . The sun and [({\color{red} \#32}){\color{blue} sky}]({\color{red} \#32}) became dark because of [(\#7)the smoke](\#7) from [(\#10)the hole](\#10) . Then [({\color{red} \#6})locusts]({\color{red} \#6}) came out of [(\#7)the smoke](\#7) and went down to [(\#8)the earth](\#8) . [({\color{red} \#6})They]({\color{red} \#6}) were given the power to sting like scorpions . [({\color{red} \#6})They]({\color{red} \#6}) were told not to damage the fields of grass or any plant or tree . [({\color{red} \#6})They]({\color{red} \#6}) were to hurt [(\#28){\color{blue} only those}](\#28) who did not have [(\#20)God 's](\#20) mark on [(\#28)their](\#28) foreheads . [({\color{red} \#6})They]({\color{red} \#6}) were not given the power to kill [(\#28)them](\#28) but only to cause [(\#28)them](\#28) pain for five months -- pain like a person feels when stung by a scorpion . During those days [(\#11)people](\#11) will look for [(\#25)a way to die](\#25) , but [(\#11)they](\#11) will not find [(\#25)it](\#25) . [(\#11)They](\#11) will want to die , but death will hide from [(\#11)them](\#11) . [({\color{red} \#6})The locusts]({\color{red} \#6}) looked like horses prepared for battle . On [({\color{red} \#6})their]({\color{red} \#6}) heads [({\color{red} \#6})they]({\color{red} \#6}) wore something that looked like a gold crown . [({\color{red} \#6})Their]({\color{red} \#6}) faces looked like [({\color{red} \#6}){\color{blue} human faces}]({\color{red} \#6}). [({\color{red} \#6})Their]({\color{red} \#6}) hair was like women 's hair . [({\color{red} \#6})Their]({\color{red} \#6}) teeth were like lions ' teeth . [({\color{red} \#6})Their]({\color{red} \#6}) chests looked like iron breastplates . The sound [({\color{red} \#6})their]({\color{red} \#6}) wings made was like the noise of many horses and chariots hurrying into battle . [({\color{red} \#6})The locusts]({\color{red} \#6}) had [({\color{red} \#6}){\color{blue} tails with stingers like scorpions}]({\color{red} \#6}) . The power [({\color{red} \#6})they]({\color{red} \#6}) had to give people pain for five months was in [({\color{red} \#6})[({\color{red} \#6})their]({\color{red} \#6}) tails]({\color{red} \#6}) . [({\color{red} \#6})They]({\color{red} \#6}) had [(\#1)a ruler , who was the angel of [(\#15)the bottomless pit](\#1)](\#15) . [(\#1)His](\#1) name in Hebrew is Abaddon . In Greek it is Apollyon . [(\#12){\color{blue} The sixth angel}](\#12) blew [(\#19)his trumpet](\#19) . Then [(\#21)I](\#21) heard [(\#22)a voice](\#22) coming from the horns on the four corners of the golden altar that is before [(\#20)God](\#20) . [(\#22)It](\#22) said to [(\#12)the sixth angel](\#12) who had [(\#19)the trumpet](\#19) , `` Free [(\#26)the four angels who are tied at the great river Euphrates](\#26) . '' [(\#26)These four angels](\#26) had been kept ready for this hour and day and month and year . [(\#26)The angels](\#26) were set free to kill [(\#24)a third of [(\#30)all the people on [(\#8)the earth](\#8)](\#24)](\#30) . [(\#21)I](\#21) heard how many troops on [(\#29)horses](\#29) were in [(\#26)their](\#26) army . In [(\#21)my](\#21) vision , [(\#21)I](\#21) saw [(\#29)the horses](\#29) and [({\color{red} \#31})the riders on [({\color{red} \#31})the horses]({\color{red} \#31})]({\color{red} \#31}) . [({\color{red} \#31})They]({\color{red} \#31}) looked like this : [({\color{red} \#31})They]({\color{red} \#31}) had breastplates that were fiery red , dark blue , and yellow like sulfur . The heads of [(\#29)the horses](\#29) looked like heads of lions . [(\#29)The horses](\#29) had [(\#14)fire](\#14) , [(\#16)[(\#4){\color{blue} smoke}](\#4) {\color{blue} , and} [(\#9){\color{blue} sulfur}](\#9)]({\color{blue} \#16}) coming out of [(\#29)[(\#17)their](\#29) mouths](\#17) . [(\#24)A third of [(\#30)all the people on earth](\#24)](\#30) were killed by these three plagues coming out of [(\#29)[(\#17)the horses '](\#29) mouths](\#17) : [(\#16)[(\#14){\color{blue} the fire}](\#14) , [(\#4){\color{blue} the smoke}](\#4) {\color{blue} , and} [(\#9){\color{blue} the sulfur}](\#9)](\#16) . [(\#29)The horses '](\#29) power was in [(\#29)[(\#17)their](\#29) mouths](\#17) and also in [(\#29)their](\#29) tails . Their tails were like snakes that have heads to bite and hurt people . [({\color{red} \#18})The other people on [(\#8)earth](\#8)]({\color{red} \#18}) were not killed by [(\#16)these plagues](\#16) . But [({\color{red} \#18})these people]({\color{red} \#18}) still did not change [({\color{red} \#18})[({\color{red} \#18})their]({\color{red} \#18}) hearts]({\color{red} \#18}) and turn away from worshiping the things [({\color{red} \#18})they]({\color{red} \#18}) had made with [({\color{red} \#18})their]({\color{red} \#18}) own hands . [({\color{red} \#18})They]({\color{red} \#18}) did not stop worshiping [(\#2){\color{blue} demons and idols}](\#2) made of gold , silver , bronze , stone , and wood -- [(\#2){\color{blue} things}](\#2) that can not see or hear or walk . [({\color{red} \#18})They]({\color{red} \#18}) did not change [({\color{red} \#18})[({\color{red} \#18})their]({\color{red} \#18}) hearts]({\color{red} \#18}) and turn away from killing other people or from [({\color{red} \#18})their]({\color{red} \#18}) evil magic , [({\color{red} \#18})their]({\color{red} \#18}) sexual sins , and [({\color{red} \#18})their]({\color{red} \#18}) stealing .

\textbf{ImCoref-CeS$_{\text{gpt4}}$ Output:} 

[(\#3)The fifth angel](\#3) blew [(\#3)his](\#3) trumpet . Then [(\#21)I](\#21) saw [(\#27)a star](\#27) fall from [({\color{red} \#32}){\color{blue} the sky}]({\color{red} \#32}) to [(\#8)the earth](\#8) . [(\#27)The star](\#27) was given [({\color{red} \#32}){\color{blue} the sky}]({\color{red} \#32}) to [(\#10)the deep hole that leads down to [(\#15)the bottomless pit](\#10)](\#15) . Then [(\#27)the star](\#27) opened [(\#10)the hole leading to [(\#15)the pit](\#10)](\#15) . [(\#7)Smoke](\#7) came up from [(\#10)the hole](\#10) like smoke from a big furnace . The sun and [({\color{red} \#32}){\color{blue} sky}]({\color{red} \#32}) became dark because of [(\#7){\color{blue} the smoke}](\#7) from [(\#10)the hole](\#10) . Then [(\#6)locusts](\#6) came out of [(\#7)the smoke](\#7) and went down to [(\#8)the earth](\#8) . [(\#6)They](\#6) were given the power to sting like scorpions . [(\#6)They](\#6) were told not to damage the fields of grass or any plant or tree . [(\#6)They](\#6) were to hurt only those who did not have [(\#20)God 's](\#20) mark on [(\#28)their](\#28) foreheads . [(\#6)They](\#6) were not given the power to kill [(\#28)them](\#28) but only to cause [(\#28)them](\#28) pain for five months -- pain like a person feels when stung by a scorpion . During those days [(\#11)people](\#11) will look for [(\#25)a way to die](\#25) , but [(\#11)they](\#11) will not find [(\#25)it](\#25) . [(\#11)They](\#11) will want to die , but death will hide from [(\#11)them](\#11) . [(\#6)The locusts](\#6) looked like horses prepared for battle . On [(\#6)their](\#6) heads [(\#6)they](\#6) wore something that looked like a gold crown . [(\#6)Their](\#6) faces looked like human faces . [(\#6)Their](\#6) hair was like women 's hair . [(\#6)Their](\#6) teeth were like lions ' teeth . [(\#6)Their](\#6) chests looked like iron breastplates . The sound [(\#6)their](\#6) wings made was like the noise of many horses and chariots hurrying into battle . [(\#6)The locusts](\#6) had tails with stingers like scorpions . The power [(\#6)they](\#6) had to give people pain for five months was in [(\#6)[(\#5)their](\#6) tails](\#5) . [(\#6)They](\#6) had [(\#1)a ruler , who was the angel of [(\#15)the bottomless pit](\#1)](\#15) . [(\#1)His](\#1) name in Hebrew is Abaddon . In Greek it is Apollyon . [(\#12){\color{blue} The sixth angel}](\#12) blew [(\#19)his trumpet](\#19) . Then [(\#21)I](\#21) heard [(\#22)a voice](\#22) coming from the horns on the four corners of the golden altar that is before [(\#20)God](\#20) . [(\#22)It](\#22) said to [(\#12)the sixth angel](\#12) who had [(\#19)the trumpet](\#19) , `` Free [(\#26)the four angels who are tied at the great river Euphrates](\#26) . '' [(\#26)These four angels](\#26) had been kept ready for this hour and day and month and year . [(\#26)The angels](\#26) were set free to kill [(\#24)a third of [(\#30)all the people on [(\#8)the earth](\#8)](\#24)](\#30) . [(\#21)I](\#21) heard how many troops on [(\#29)horses](\#29) were in [(\#26)their](\#26) army . In [(\#21)my](\#21) vision , [(\#21)I](\#21) saw [(\#29)the horses](\#29) and [(\#31)the riders on [({\color{red} \#34})the horses]({\color{red} {\color{red} \#34}})](\#31) . [(\#31)They](\#31) looked like this : [(\#31)They](\#31) had breastplates that were fiery red , dark blue , and yellow like sulfur . The heads of [(\#29)the horses](\#29) looked like heads of lions . [(\#29)The horses](\#29) had [(\#14)fire](\#14) , [(\#4)smoke](\#4) , and [(\#9)sulfur](\#9) coming out of [(\#29)[(\#17)their](\#29) mouths](\#17) . [(\#24)A third of [(\#30)all the people on earth](\#24)](\#30) were killed by these three plagues coming out of [(\#29)[(\#17)the horses '](\#29) mouths](\#17) : [(\#16)[(\#14)the fire](\#14) , [(\#4){\color{blue} the smoke}](\#4) {\color{blue} , and} [(\#9){\color{blue} the sulfur}](\#9)]({\color{blue} \#16}) . [(\#29)The horses '](\#29) power was in [(\#29)[(\#17)their](\#29) mouths](\#17) and also in [(\#29)their](\#29) tails . Their tails were like snakes that have heads to bite and hurt people . [(\#18)The other people on [(\#8)earth](\#8)](\#18) were not killed by [(\#16)these plagues](\#16) . But [(\#18)these people](\#18) still did not change [(\#18)[({\color{red} \#33})their](\#18) hearts]({\color{red} \#33}) and turn away from worshiping the things [(\#18)they](\#18) had made with [(\#18)their](\#18) own hands . [(\#18)They](\#18) did not stop worshiping [(\#2){\color{blue} demons and idols}](\#2) made of gold , silver , bronze , stone , and wood -- [(\#2){\color{blue} things}](\#2) that can not see or hear or walk . [(\#18)They](\#18) did not change [(\#18)[({\color{red} \#33})their](\#18) hearts]({\color{red} \#33}) and turn away from killing other people or from [(\#18)their](\#18) evil magic , [(\#18)their](\#18) sexual sins , and [(\#18)their](\#18) stealing .

\end{document}